\documentclass[5p,times]{elsarticle}
\usepackage[switch]{lineno}
\usepackage{hyperref}

\usepackage[table]{xcolor}

\usepackage{amsmath,amssymb,amsfonts}
\usepackage{kotex}

\usepackage{mathtools}
\DeclarePairedDelimiter\ceil{\lceil}{\rceil}

\usepackage{amsmath}
\DeclareMathOperator*{\argmax}{argmax} 
\usepackage{multirow}
\usepackage{booktabs}
\usepackage{pifont}
\usepackage{xcolor}
\newcommand{\cmark}{\textcolor{green!80!black}{\ding{51}}}
\newcommand{\xmark}{\textcolor{red}{\ding{55}}}
\newcommand{\pmark}{\textcolor{blue}{\ding{115}}}

\usepackage{algorithm}
\usepackage{algpseudocode}

\usepackage{graphicx}
\usepackage{textcomp}
\usepackage[caption=false,labelfont=sf,textfont=sf]{subfig} 

\definecolor{mypink}{rgb}{0,0,0}
\newcommand{\news}[1]{\textcolor{mypink}{#1}} 

\makeatletter
\def\therule{\makebox[\algorithmicindent][l]{\hspace*{.5em}\vrule height .75\baselineskip depth .25\baselineskip}}%

\newtoks\therules
\therules={}
\def\appendto#1#2{\expandafter#1\expandafter{\the#1#2}}
\def\gobblefirst#1{
  #1\expandafter\expandafter\expandafter{\expandafter\@gobble\the#1}}%
\def\LState{\State\unskip\the\therules}
\def\pushindent{\appendto\therules\therule}%
\def\popindent{\gobblefirst\therules}%
\def\printindent{\unskip\the\therules}%
\def\printandpush{\printindent\pushindent}%
\def\popandprint{\popindent\printindent}%

\algdef{SE}[WHILE]{While}{EndWhile}[1]
  {\printandpush\algorithmicwhile\ #1\ \algorithmicdo}
  {\popandprint\algorithmicend\ \algorithmicwhile}%
\algdef{SE}[FOR]{For}{EndFor}[1]
  {\printandpush\algorithmicfor\ #1\ \algorithmicdo}
  {\popandprint\algorithmicend\ \algorithmicfor}%
\algdef{S}[FOR]{ForAll}[1]
  {\printindent\algorithmicforall\ #1\ \algorithmicdo}%
\algdef{SE}[LOOP]{Loop}{EndLoop}
  {\printandpush\algorithmicloop}
  {\popandprint\algorithmicend\ \algorithmicloop}%
\algdef{SE}[REPEAT]{Repeat}{Until}
  {\printandpush\algorithmicrepeat}[1]
  {\popandprint\algorithmicuntil\ #1}%
\algdef{SE}[IF]{If}{EndIf}[1]
  {\printandpush\algorithmicif\ #1\ \algorithmicthen}
  {\popandprint\algorithmicend\ \algorithmicif}%
\algdef{C}[IF]{IF}{ElsIf}[1]
  {\popandprint\pushindent\algorithmicelse\ \algorithmicif\ #1\ \algorithmicthen}%
\algdef{Ce}[ELSE]{IF}{Else}{EndIf}
  {\popandprint\pushindent\algorithmicelse}%
\algdef{SE}[PROCEDURE]{Procedure}{EndProcedure}[2]
   {\printandpush\algorithmicprocedure\ \textproc{#1}\ifthenelse{\equal{#2}{}}{}{(#2)}}%
   {\popandprint\algorithmicend\ \algorithmicprocedure}%
\algdef{SE}[FUNCTION]{Function}{EndFunction}[2]
   {\printandpush\algorithmicfunction\ \textproc{#1}\ifthenelse{\equal{#2}{}}{}{(#2)}}%
   {\popandprint\algorithmicend\ \algorithmicfunction}%
\makeatother

\modulolinenumbers[5]

\journal{Future Generation Computer Systems}

\biboptions{sort&compress}

\begin{document}

\begin{frontmatter}

\title{Quantune: Post-training Quantization of Convolutional Neural Networks using Extreme Gradient Boosting for Fast Deployment}
\author{Jemin Lee\corref{mycorrespondingauthor}}
\cortext[mycorrespondingauthor]{Corresponding author}
\ead{leejaymin@etri.re.kr}

\author{Misun Yu}
\ead{msyu@etri.re.kr}

\author{Yongin Kwon}
\ead{yongin.kwon@etri.re.kr}

\author{Taeho Kim}
\ead{taehokim@etri.re.kr}

\address{Artificial Intelligence Research Laboratory, Electronics and Telecommunications Research Institute (ETRI), Daejeon 34129, South Korea}

\begin{abstract}
To adopt convolutional neural networks (CNN) for a range of resource-constrained targets, it is necessary to compress the CNN models by performing quantization, whereby precision representation is converted to a lower bit representation. To overcome problems such as sensitivity of the training dataset, high computational requirements, and large time consumption, post-training quantization methods that do not require retraining have been proposed. In addition, to compensate for the accuracy drop without retraining, previous studies on post-training quantization have proposed several complementary methods: calibration, schemes, clipping, granularity, and mixed-precision. To generate a quantized model with minimal error, it is necessary to study all possible combinations of the methods because each of them is complementary and the CNN models have different characteristics. However, an exhaustive or a heuristic search is either too time-consuming or suboptimal. To overcome this challenge, we propose an auto-tuner known as \textsf{Quantune}, which builds a gradient tree boosting model to accelerate the search for the configurations of quantization and reduce the quantization error. \news{We evaluate and compare \textsf{Quantune} with the \textit{random}, \textit{grid}, and \textit{genetic} algorithms.} The experimental results show that \textsf{Quantune} reduces the search time for quantization by approximately \news{$36.5\times$} with an accuracy loss of 0.07–0.65\% across six CNN models, including the fragile ones (MobileNet, SqueezeNet, and ShuffleNet). To support multiple targets and adopt continuously evolving quantization works, \textsf{Quantune} is implemented on a full-fledged compiler for deep learning as an open-sourced project.
\end{abstract}


\end{frontmatter}


\section{Introduction}
\label{sec:introduction}
In spite of the prevalence of convolutional neural networks (CNNs), the high computational requirements restrict their use in resource-constrained devices~\cite{astrid2018deep}. To address this challenge, considerable research is being done on the compression of neural networks. One of the promising compression methods is quantization, which converts high-precision representations (32-bit floating point) into lower bit representations (\texttt{int8} fixed point). 
Quantization can reduce the model size of the CNN, memory footprint, and energy consumption and improve the inference time by utilizing special instructions supported by the hardware platforms.

To compensate for the accuracy drop of the quantized models, most of the quantization methods consider retraining~\cite{kris_whitepaper2018,steven2020,PACT2018,zhang2018lq,jung2019learning,zhou2016dorefa,jacob2018,songhan2016}. However, this method, which is commonly referred to as quantization-aware training (hereinafter referred to as QAT), is not widely adopted in real-world scenarios because of the following issues.
First, a full-size dataset is often unavailable owing to privacy concerns or because it is proprietary information. Second, the retraining process in QAT is time-consuming and resource-hungry because of the long periods of tuning. Third, its hyper-parameter tuning is complicated because it requires considerable expertise to develop the architecture of CNNs. Such limitations prevent us from deploying quantized models in a timely manner.

In practice, post-training quantization (hereinafter referred to as PTQ) methods are widely utilized owing to their good applicability~\cite{kris_whitepaper2018,jiang2021automated,banner_neurips2019,choukroun2019low,zhao2019improving,lee2018quantization,goncharenko2019fast,migacz20178,wu2020integer}. 
To recover the accuracy drop, previous studies on PTQ have proposed the following diverse complementary methods: novel schemes for mapping~\cite{kris_whitepaper2018,jiang2021automated}, calibration for activation of quantization~\cite{kris_whitepaper2018,jiang2021automated}, granularity for sharing quantization parameters among tensor elements~\cite{kris_whitepaper2018,jiang2021automated}, clipping~\cite{banner_neurips2019,choukroun2019low,zhao2019improving,goncharenko2019fast,migacz20178}, and mixed-precision~\cite{wu2020integer}.
In this study, we define \textit{a quantization configuration} as a combination of such complementary methods. Based on our experiments, as shown in Table~\ref{table:toprnk}, the quantization configurations vary with the target CNN models to attain optimal results. Nonetheless, possible quantization configurations vary depending on the target hardware devices. For example, complicated quantifiers are not tolerated on highly constrained hardware like integer-only accelerators~\cite{moreau2018leveraging,zhao2019linear}.
Therefore, it is necessary to perform the configuration search every time depending on the model and the hardware across all the combinations. However, it is daunting to explore all the possible configurations when quantization requests occur.

To overcome this challenge, we propose an auto-tuner for configuration search using a gradient tree boosting model (based on XGBoost~\cite{chen2016xgboost}) known as \textsf{Quantune}. \textsf{Quantune} generates quantized models without noticeable accuracy degradation and retraining and supports multiple hardware platforms (CPU, GPU, and integer-only accelerator) by implementing them on an open-source deep learning compiler stack. 
We assume that the features extracted from the CNN models are related to quantization configurations. 
Based on such an assumption, we build an XGBoost model, considering the features, configurations, and accuracy. 
We perform our search algorithm to find the quantization configuration. 
To support multiple targets as a unified quantization model format, we implement \textsf{Quantune} on a compiler stack known as Glow. For a rich configuration, we extend Glow to enable layer-wise mixed precision (\textit{int8} and \textit{fp32}) and integer-only quantization. \textsf{Quantune} enables the generation of quantized models for a range of targets including the CPU, GPU, and integer-only hardware such as an accelerator. For the integer-only hardware, we use Versatile Tensor Accelerator (VTA)~\cite{moreau2019hardware} as an open-hardware architecture. To support the VTA, the entire inference processes comprise only integer multiplication, addition, and bit-shifting.

We evaluate \textsf{Quantune}, in terms of the efficiency of the proposed search algorithm, accuracy of the quantized models, and end-to-end latency of the embedded CPU, server-side CPU, and GPU. 
\news{The experimental results show that \textsf{Quantune} is 1.3-36.5 times faster than the \textit{random}, \textit{grid}, and \textit{genetic} algorithms and achieves better quality in terms of quantization for the six models across the CPU, GPU and NPU.} To prove the quality of the quantized models, \textsf{Quantune} is compared with mature tools such as TensorRT and TVM, which support the PTQ on the GPU and integer-only accelerator (VTA), respectively. Regarding TensorRT, the experimental results show competitive accuracy of the quantized models in ResNet18, ResNet50, ShuffleNet, and SqueezeNet, and better results in GoogleNet.  \textsf{Quantune} precisely quantizes fragile models such as MobileNet, ShuffleNet, and SqueezeNet despite their small representational capacities~\cite{jacob2018quantization}. 
In the case of the integer-only accelerator, \textsf{Quantune} achieves a 32.52\% improvement in accuracy compared to TVM-VTA~\cite{moreau2018leveraging}.

By implementing \textsf{Quantune} on the compiler stack, the quantized models can be performed on the CPU, GPU, and the accelerator. Therefore, the executable binaries for the CPU, GPU, and accelerator are generated from the quantized models. We measure the end-to-end inference time of the quantized models on the edge-side CPU, server-side CPU, and server-side GPU. \textsf{Quantune} achieves speedups of 0.34–1.22, 0.27–2.6, and 0.93–1.57 on ARM-A53, Intel-i7-8700, and NVIDIA 2080ti GPU, respectively, against \textit{fp32} execution. From the experiment, it can be seen that latency is not improved for all the quantized models because the extended compiler does not exploit fast 8-bit multiply-accumulate instruction (ARM-vmlal, Intel VNNI, and NVIDIA-DP4A) provided by hardware vendors while generating the kernel code. This indicates that quantization is not only a method that can reduce the memory footprint as in conventional applications, but also a mandatory step to develop deployable CNN models on the integer-only accelerator. The efficient kernel code generation for the quantized models is an important research direction; however, that is beyond the scope of this paper.

The main contributions of our work are summarized as follows.
\begin{itemize}

\item We show that the optimal configurations for quantization are diverse depending on the CNN models. To demonstrate the diversity of the quantization configurations, we conduct entropy analysis. As a result, it is found that the entropy of each complementary method is not the same across all the CNN models. There is no universal configuration that is always applicable regardless of the type of CNN models.

\item To efficiently explore all combinations of the quantization model, we propose \textsf{Quantune}, which combines both XGBoost and transfer learning to seek the optimal configuration. \news{\textsf{Quantune} significantly outperforms the \textit{grid}, \textit{random}, and \textit{genetic} algorithms by approximately 36.5$\times$ with a 0.07-0.65 accuracy loss across the six CNN models.}

\item For practical use and as an extension, \textsf{Quantune} was implemented on the open-source compiler stack known as Glow~\cite{rotem2018glow}, instead of performing a pure algorithm design. The extended Glow provides layer-wise mixed precision and integer-only quantization. Therefore, we generated the binary code of the quantized models for diverse hardware targets ranging from CPU (x86 and ARM) to the integer-only accelerator (VTA). To support the integer-only accelerator, \textsf{Quantum} not only quantized weights, activations, bias, and scales, but also generated a computational graph that is comprised of integer multiplication, addition, and bit shift without any floating-point computation.

\item Regarding the quality of the quantized models, \textsf{Quantune} achieves 0.59\% better accuracy in GoogleNet slim v4 than TensorRT-7.2.2 on the NVIDIA GPU. Regarding the integer-only quantization, \textsf{Quantune} significantly outperforms the previous result (based on single-scale quantization across the whole layer) by approximately 32.52\%. In addition, we directly measure the end-to-end inference time of the quantized models on a real CPU and GPU.

\end{itemize}

\begin{table*}[t]
\centering
\caption{The best results among all the possible quantized models for six CNN models. Hereafter, we abbreviate MobileNet V2 as "MN", ShuffleNet V1 as "SHN", SqueezeNet V1 as "SQN", GoogleNet Slim V4 as "GN", ResNet18 V1 as "RN18", and ResNet50 V1 as "RN50".}
\label{table:toprnk}
\resizebox{\textwidth}{!}{
\begin{tabular}{lrrrrrr
} 
\toprule
\multicolumn{1}{c}{\textbf{Model Name}} & \multicolumn{1}{c}{\textbf{Precision}} & \multicolumn{1}{c}{\textbf{\# of Images for Calibration}} & \multicolumn{1}{c}{\textbf{Granularity}} & \multicolumn{1}{c}{\textbf{Clipping}} & \multicolumn{1}{c}{\textbf{Scheme}} & \multicolumn{1}{c}{\textbf{Accuracy (Error)}} 
\\   \midrule
MobileNet V2(MN) & \texttt{int8} & 1,000 & Channel  & KL  & Asymmetric  & $71.23(\textcolor{red}{-0.58})\%$ \\
ShuffleNet V1(SHN) & \texttt{int8}+\texttt{fp32} & 1 & Channel  & Max  & Symmetric uint & $63.59(\textcolor{red}{-0.37})\%$ \\ 
SqueezeNet V1(SQN) & \texttt{int8} & 1,000 & Channel  & KL  & Asymmetric & $53.15(\textcolor{red}{-0.65})\%$ \\ 
GoogleNet Slim V4(GN) & \texttt{int8} & 1,000  & Tensor & KL & Asymmetric & $70.58(\textcolor{blue}{+0.19})\%$ \\ 
ResNet18 V1(RN18) & \texttt{int8} & 1,000 & Tensor & KL  & Asymmetric  & $70.25(\textcolor{red}{-0.42})\%$ \\ 
ResNet50 V1(RN50) & \texttt{int8} & 10,000 & Channel  & KL  & Asymmetric  & $76.01(\textcolor{red}{-0.07})\%$\\ \bottomrule
\end{tabular}
}
\end{table*}

\section{Related Work} 
Quantization has attracted significant attention owing to its tangible benefits for model compression. In this section, we categorize previous studies on quantization into post-training quantization and quantization-aware training and describe the novelty of our study in each category by comparing it to the existing tools.

\subsection{Quantization-aware Training}
Quantization-aware training (QAT) methods map high bit precision to low bit precision using training step~\cite{kris_whitepaper2018,steven2020,PACT2018,zhang2018lq,jung2019learning,zhou2016dorefa,jacob2018,songhan2016}.
QAT reduces the accuracy drop from the quantized model by using a retraining procedure that is performed for a few epochs.
Owing to retraining, QAT is able to quantize CNN models in low precision representation without noticeable accuracy drop and can even operate at 2 bits.
\news{However, QAT has the following limitations: (i) retraining is time-consuming, (ii) the training data are not always accessible by third party services, and (iii) its hyper-parameter tuning is  complicated.
Even using active and continual learning works does not completely mitigate these limitations~\cite{nguyen2017variational,doulamis2000line,shin2017fixed,dhaliwal2018effective}.}
Therefore, considering its rapid deployment and practical usage, we focus on post-training quantization that does not require retraining based on the training data.

\subsection{Post-training Quantization}
Post-training quantization (hereafter called PTQ) methods map high precision representation bits to low-precision bits without re-training steps~\cite{kris_whitepaper2018,jiang2021automated,banner_neurips2019,choukroun2019low,nagel2019data,zhao2019improving,lee2018quantization,goncharenko2019fast,meller19a,migacz20178,wu2020integer}.


Post-training quantization is widely adopted in practical cases because it is not necessary to access the full training dataset for retraining. Post-training quantization remedies the large time-consumption for retraining and the data privacy issue. 
Therefore, it helps rapid deployment of the CNN models on resource-constrained devices.
Typically, PTQ leads to non-trivial accuracy degradation, especially in low precision representations. 
Owing to the prevalence of \textit{int8} data type support in the many hardware platforms, most of the previous studies on PTQ have focused on \texttt{int8} quantization and proposed several methods such as diverse schemes, clippings, and mixed-precision to recover the accuracy drop.
However, our experimental result shows that the quantization configurations for the best accuracy are dependent on the CNN models. There is no universal configuration that is always applied to attain the most accurately quantized models. Considering each CNN model, a naive parameter search is time-consuming. We overcome this challenge to seek the best configurations for each CNN model. Empirically, this study is the first work to find the optimal quantization strategy using the machine learning model.

\subsection{Deep Learning Compilers}
The increasing demands for efficiency on the deep learning (DL) models has made deep DL compilers prevalent. Deep learning compilers have been proposed in both academia and industry. Most of the DL compilers focus on improving the quantization ability of a single class of hardware platform such as Intel nGraph~\cite{cyphers2018intel}, NVIDIA TensorRT~\cite{migacz20178}, ARM NN\footnote{https://github.com/ARM-software/armnn}, and Xilinx Vitis\footnote{https://www.xilinx.com/products/design-tools/vitis/vitis-platform.html}. 
Nonetheless, it is difficult to extend such tools to other hardware platforms owing to their proprietaries. On the contrary, there are community-driven DL compilers for multiple hardware platforms: TVM~\cite{chen2018tvm} and Glow~\cite{rotem2018glow}. 
Such open-source DL compilers support adequate capabilities to adopt diverse quantization settings on multiple targets. Each of them needs to go through manual search procedures to find the optimal quantization settings because of the lack of an auto-tuner, hindering the rapid deployment of the models. \textsf{Quantune} complements the existing studies on the open DL compilers. \textsf{Quantune} introduces a novel search algorithm that generates the optimal quantized models, considering accuracy. With the integration on Glow, \textsf{Quantune} employs full advantages of its compiler to generate kernel codes on multiple targets.

\begin{figure*}[t]
	\centering
	\includegraphics[width=1.5\columnwidth]{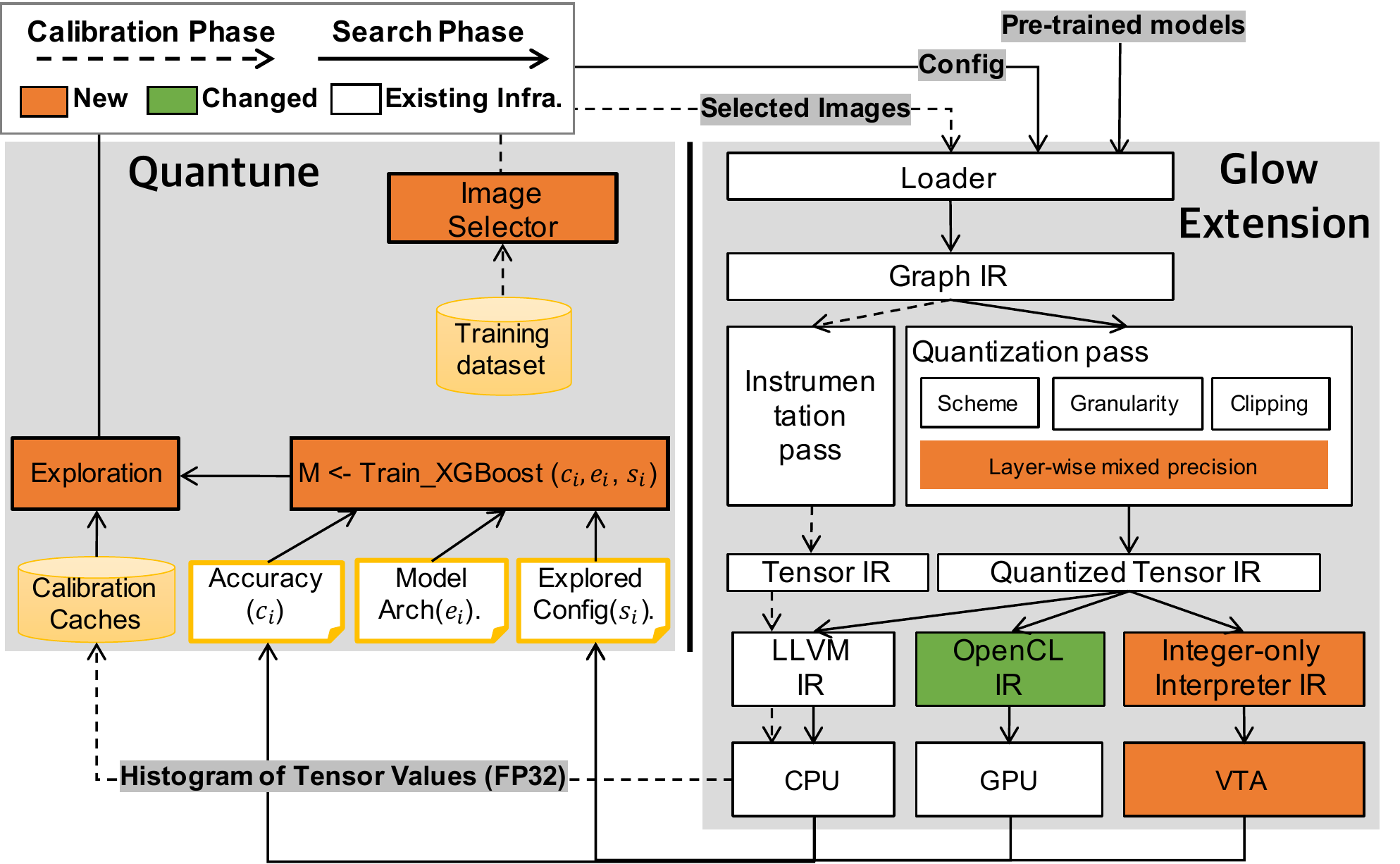}
	\caption{System overview of the proposed quantization framework. Our framework consists of two modules (\textsf{Quantune} and \textsf{Glow Extension}) and two phases (Calibration and Search).}
	\label{fig:overview}
\end{figure*}

\section{Overview}
This section describes the overall procedure for quantization by presenting each module.
Fig.~\ref{fig:overview} shows the overall workflow for the configuration search (\textsf{Quantune}) and code generation of the quantized CNN models (Glow Extension). We implement \textsf{Quantune} onto Glow~\cite{rotem2018glow} (an open-source DL compiler). 
Therefore, our quantization method can support a variety of target devices and bring about rapid deployment. In addition, we release the code of Glow extension as a part of NEST-C\footnote{https://github.com/etri/nest-compiler} and \textsf{Quantune} implemented in R\footnote{https://github.com/leejaymin/qaunt\_xgboost}. 
The two colored components in Fig.~\ref{fig:overview} represent either the changed or new ones. In the \textsf{Quantune} module part, all the components are fully developed for the configuration search. Considering the Glow, our extension aims to support the integer-only accelerator and provide layer-wise quantization for mixed-precision. The overall process of quantization goes through two phases: calibration and configuration search.

\textbf{Calibration Phase.} The dashed-lines indicate the whole procedure of the calibration phase in Fig.~\ref{fig:overview}. 
\news{In this phase, the histogram of possible numeric ranges in each layer of the neural network is captured for the activation of the quantization and saved to a calibration cache.}
\news{First, for the calibration, the Glow compiler takes a pre-trained model and an image as input.}
The images are selected from the training dataset using the Image Selector. Second, the Glow generates the instrumented codes by moving through the Loader, Graph-IR, and Tensor-IR. Concerning the Graph IR, the Glow performs two kinds of optimizations: target independent and target dependent passes. In Tensor-IR level, the Glow determines a schedule of operators while optimizing the memory usage. Finally, the histogram of the tensor values is generated by observing the execution during the inference to capture the possible numeric ranges of activations in each layer of the neural network.

\textbf{Search Phase.} The solid-lines indicate the whole procedure of the search phase in Fig.~\ref{fig:overview}. 
In this phase, the quantized models and optimized codes are generated with the configuration. As mentioned earlier, the quantization error varies in a chosen configuration. To quickly find the optimal configuration, \textsf{Quantune} efficiently explores the possible configurations based on the XGBoost models that are trained with the model architecture ($e_{i}$), explored configuration ($s_{i}$), and measured accuracy ($c_{i}$).
All the possible configurations are detailed in Section~\ref{sec:quantization_methodology}. 
Further details of the search algorithm are described in Section~\ref{sec:modeling}.

\begin{figure*}[t]
	\centering
	\subfloat[][MobileNet V2]{\label{fig:q_mobilenet}\includegraphics[width=.33\textwidth]{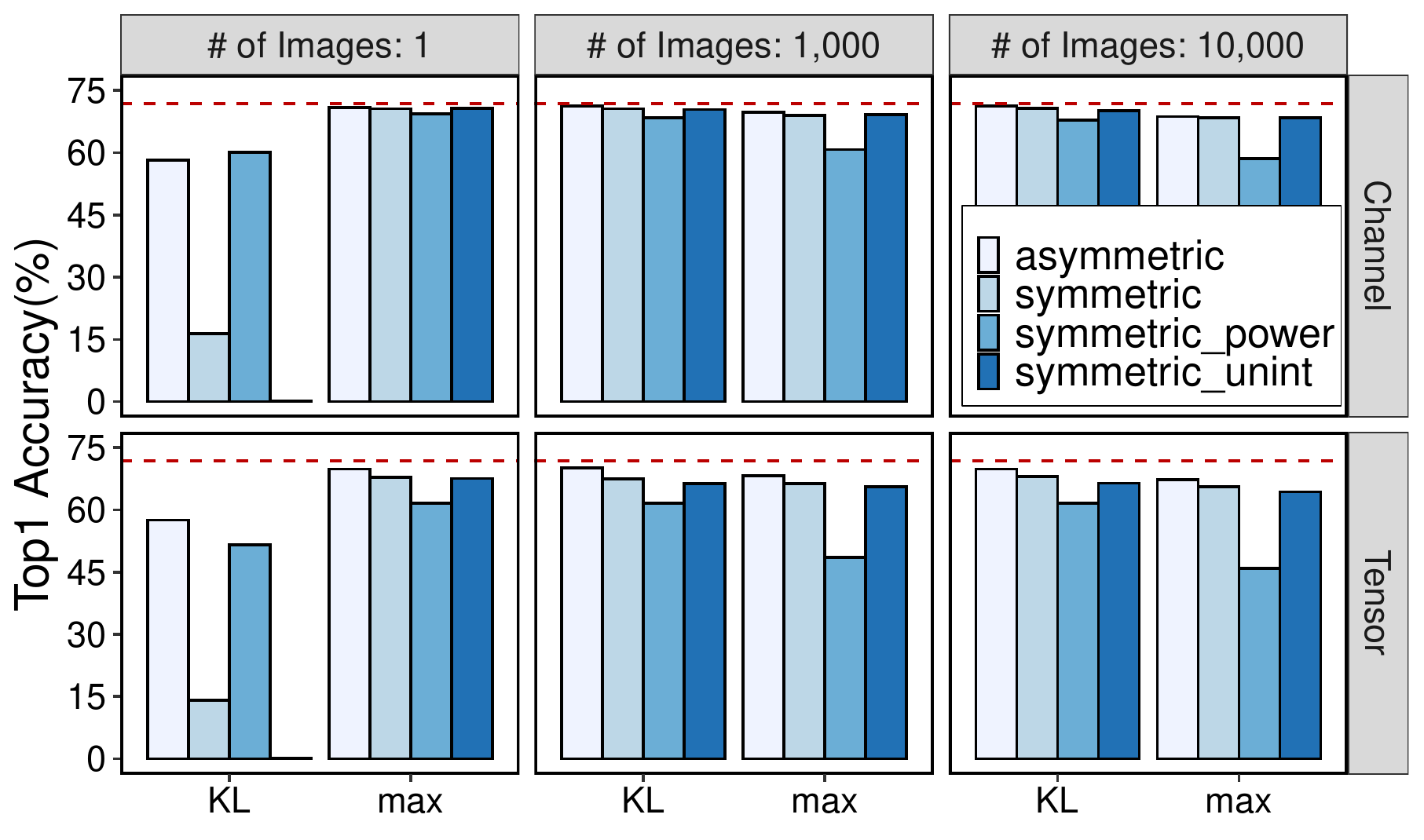}}
	\subfloat[][SqueezeNet V1]{\label{fig:q_squeezenet}\includegraphics[width=.33\textwidth]{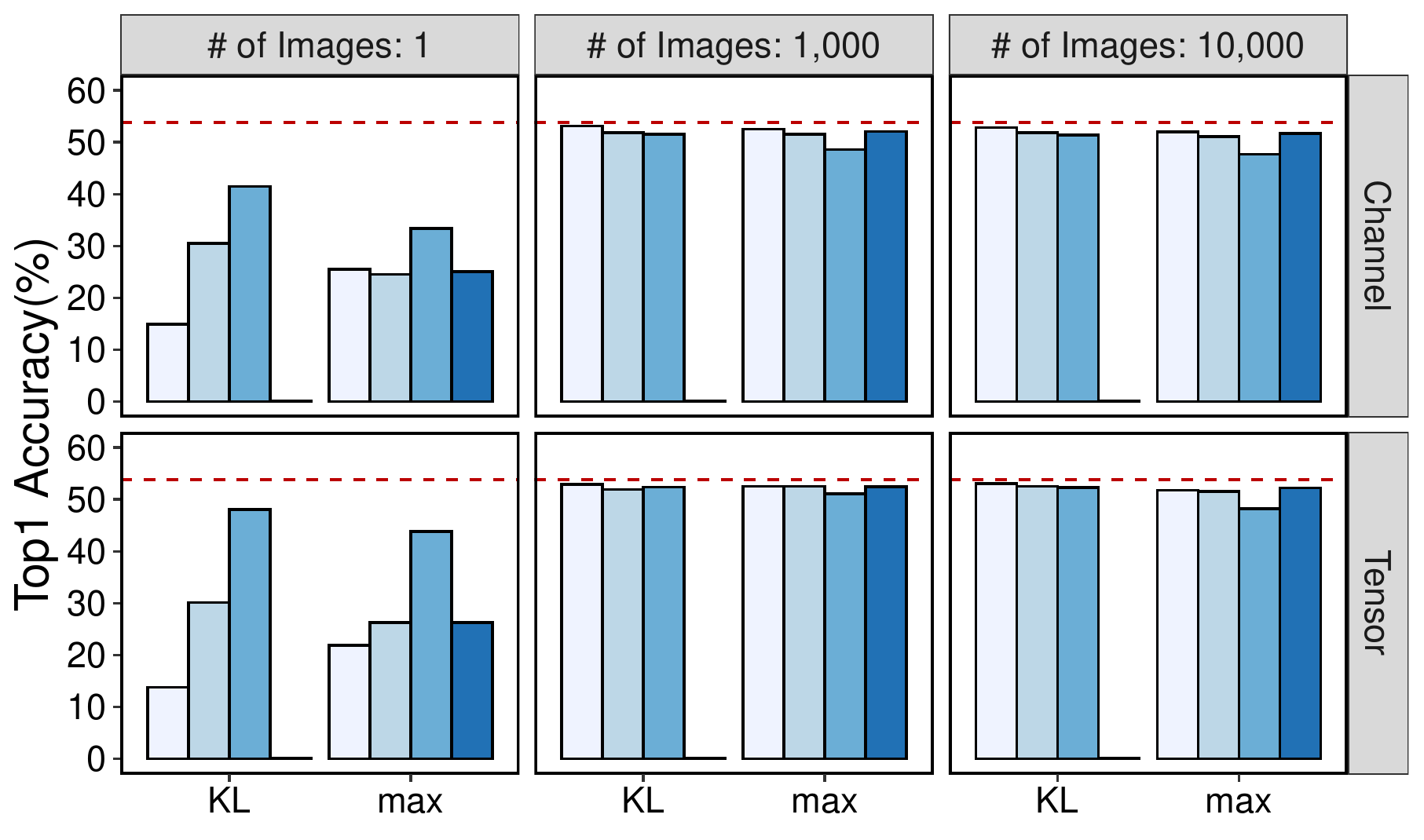}}
	\subfloat[][ResNet18 V1]{\label{fig:q_resnet18}\includegraphics[width=.33\textwidth]{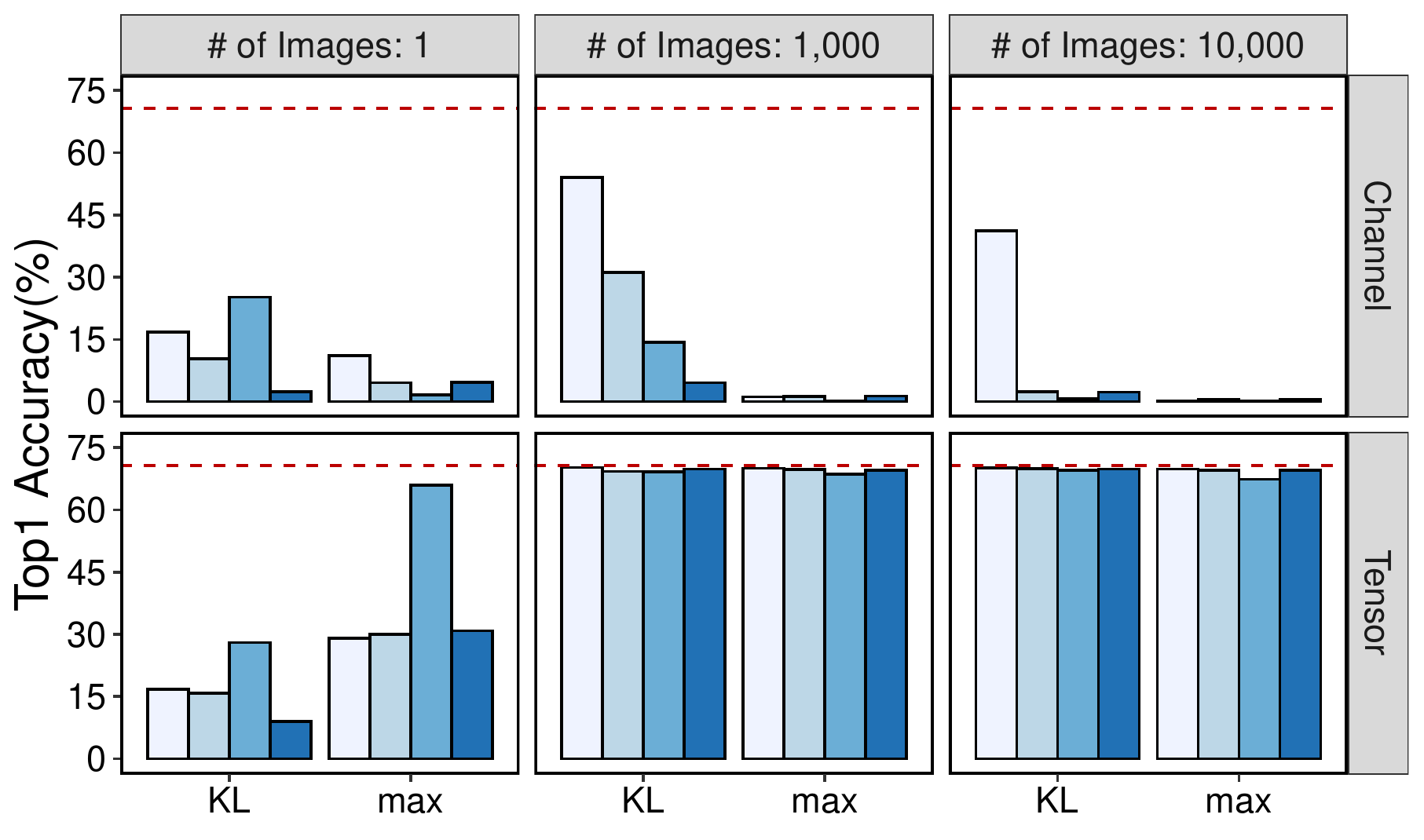}}\vfill
	\subfloat[][ResNet50 V1]{\label{fig:q_resnet50}\includegraphics[width=.33\textwidth]{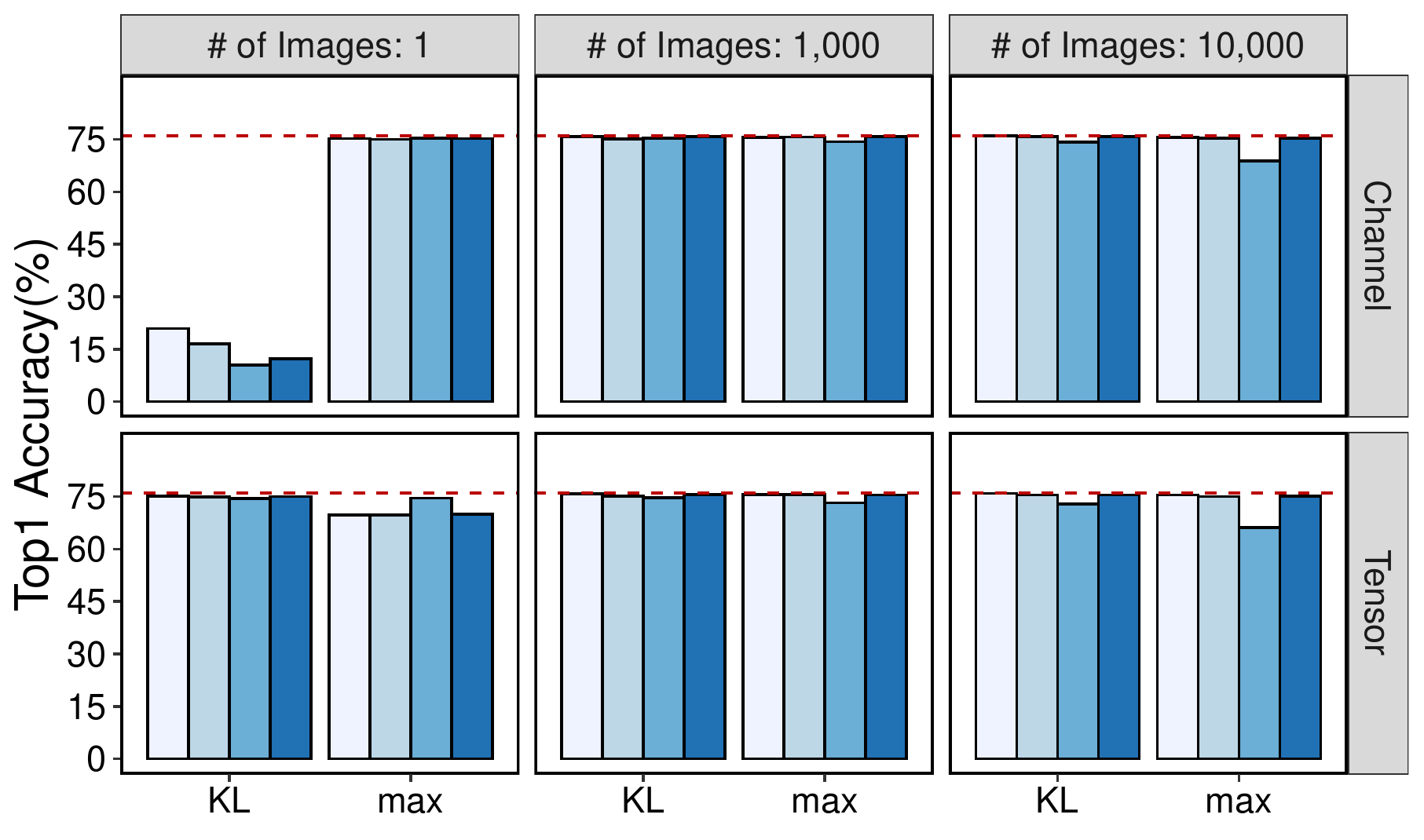}}	
	\subfloat[][ShuffleNet V1]{\label{fig:q_shufflenet}\includegraphics[width=.33\textwidth]{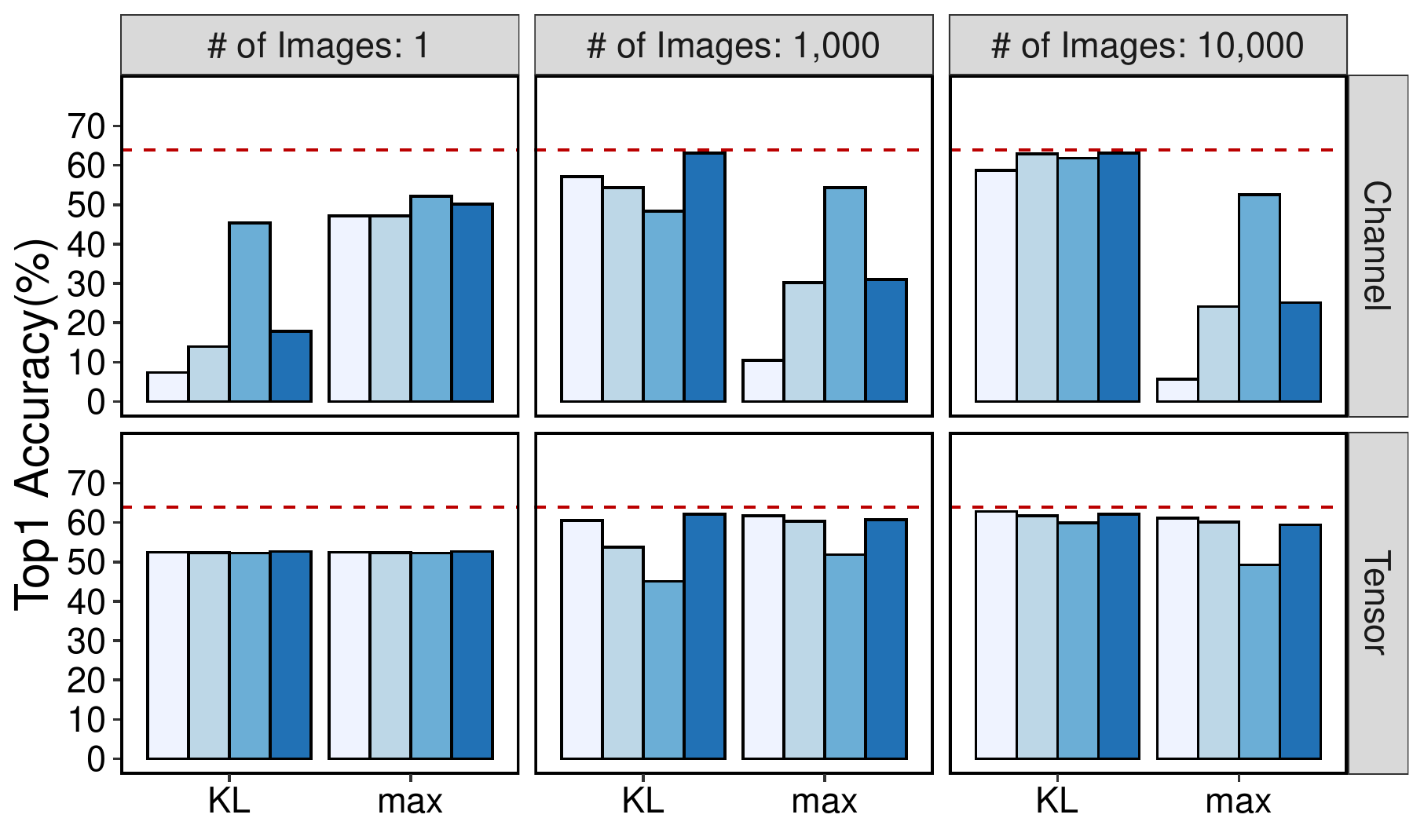}}	
	\subfloat[][GoogleNet Slim V4]{\label{fig:q_googlenet}\includegraphics[width=.33\textwidth]{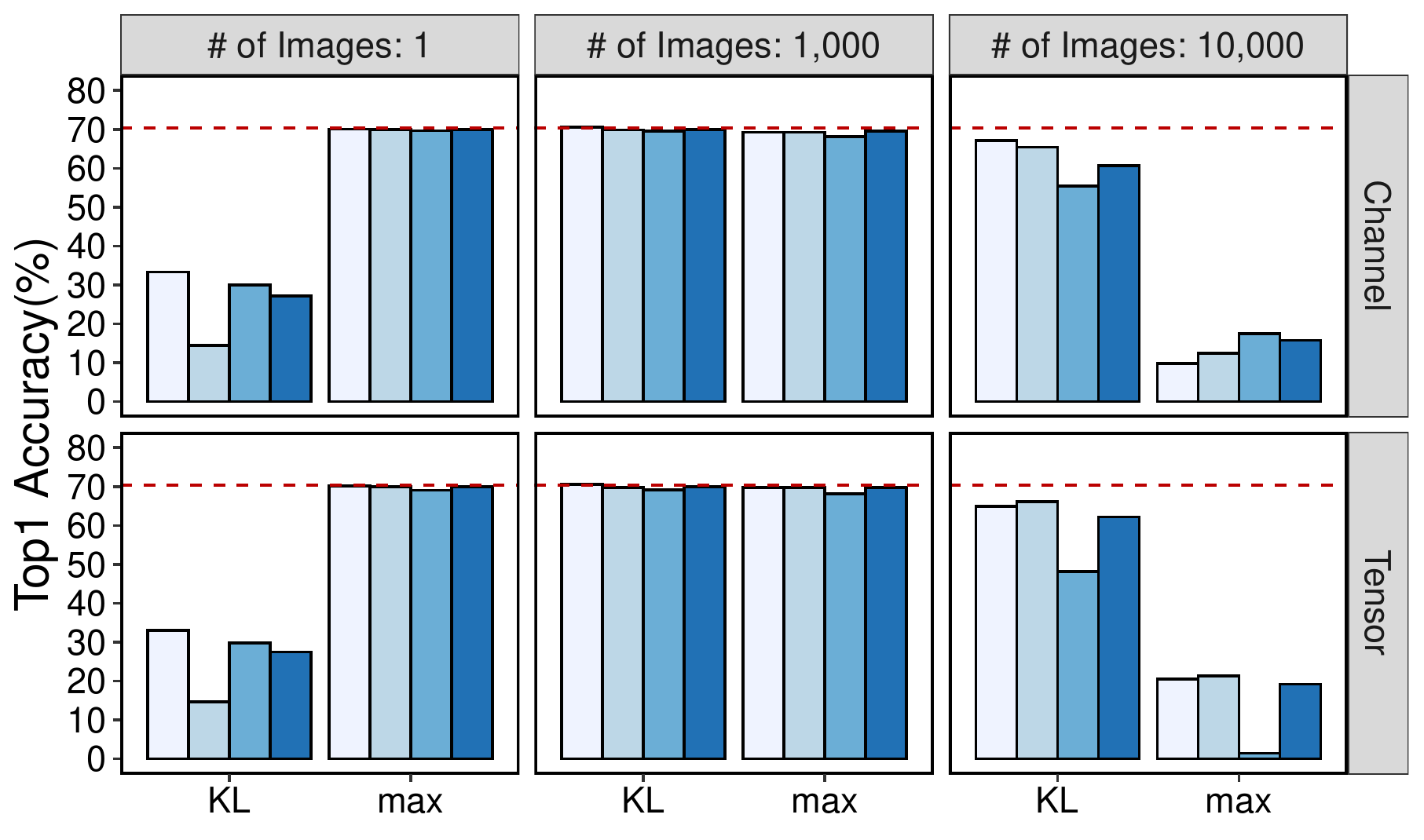}}\vfill
	\subfloat[][ShuffleNet V1(mixed precision)]{\label{fig:q_shufflenet_mixed}\includegraphics[width=.33\textwidth]{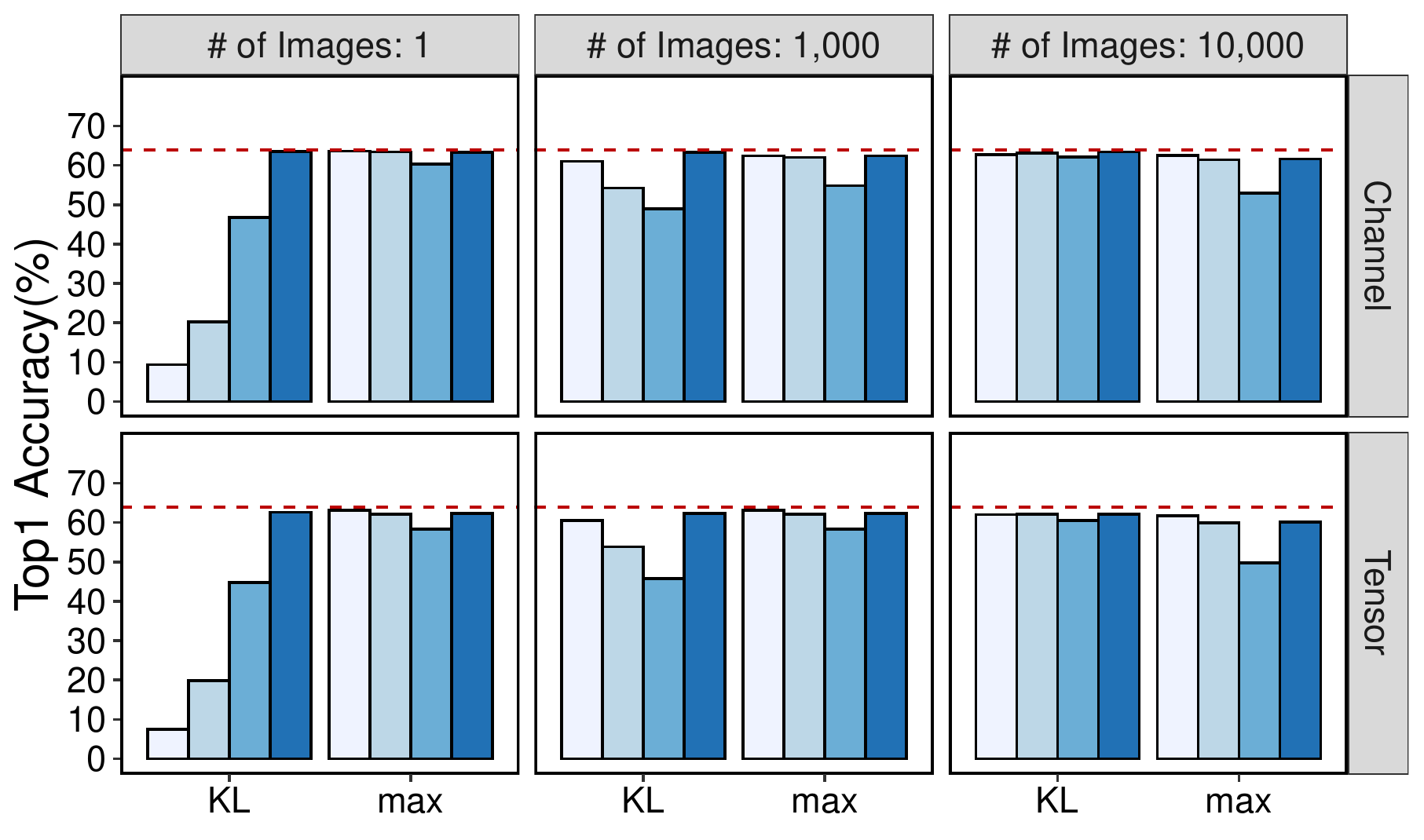}}	
	\subfloat[][GoogleNet Slim V4 (mixed precision)]{\label{fig:q_googlenet_mixed}\includegraphics[width=.33\textwidth]{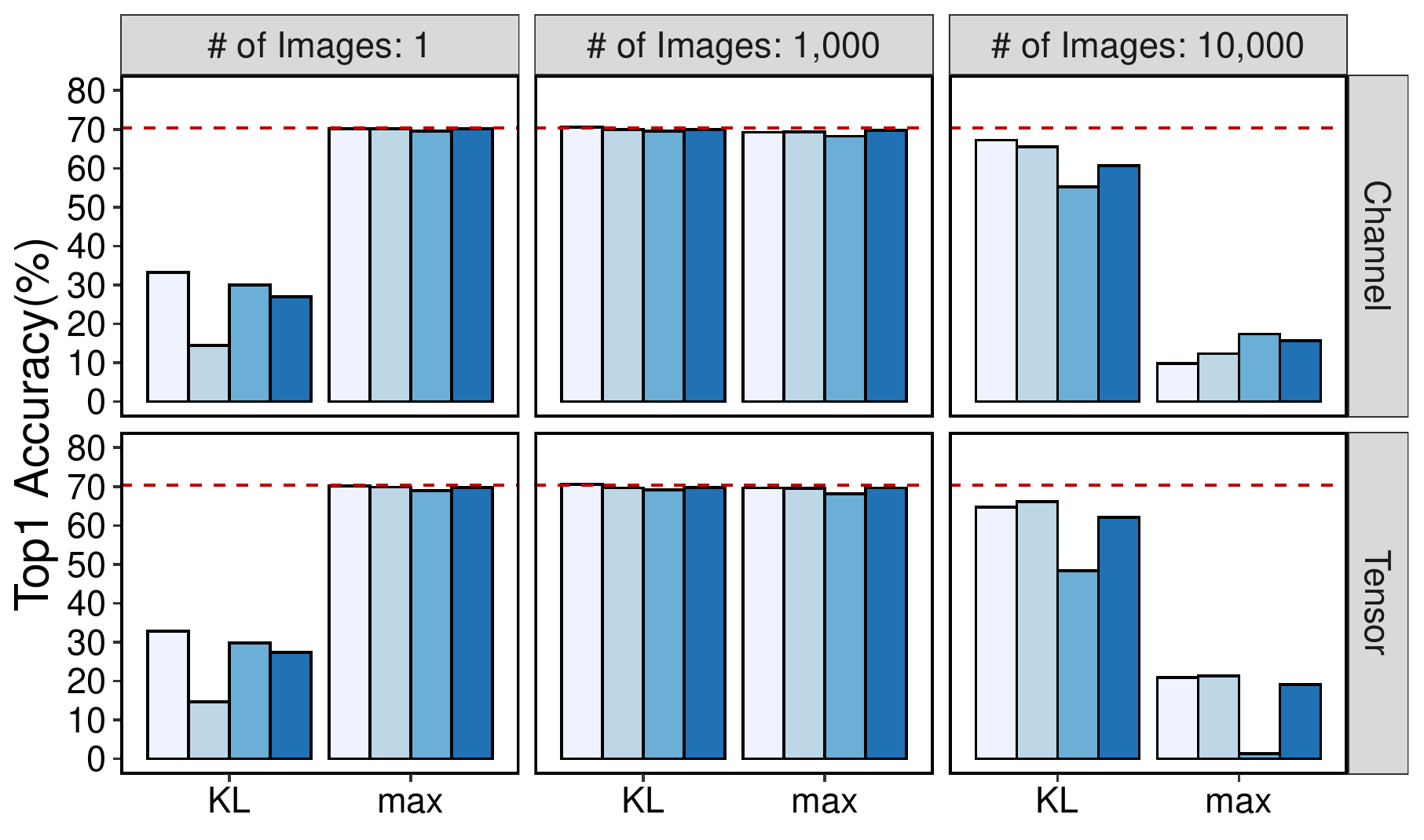}}\vfill

	\caption{Accuracy with \texttt{int8} quantization of weights and activations. The considered configurations are calibration, schemes, clipping, granularity, and mixed-precision. The dashed-lines indicate full-precision accuracy (\texttt{fp32}).}
	\label{fig:combi_qm}
\end{figure*}

\section{Quantization Methodology}\label{sec:quantization_methodology}
\news{In the calibration phase, \textsf{Quantune} collected calibration data and saved them into the calibration cache. The calibration cache containing the distribution of numerical data as a histogram was used to obtain accurate thresholds for each tensor in the original \textit{fp32} model.} 
The quantized models are generated by calculating the scale (a quantization parameter) based on the collected data of the activation tensors.
The search space of the possible configuration for quantization is the combination of five complementary methods (calibration, scheme, clipping, granularity, and mixed-precision) that affect the accuracy of the quantized models.
Equation (\ref{eq:overall_quantization}) denotes the space of the possible configurations.

\begin{multline}
Search\;Space(96)= \news{Calibration\;Cache(3)} \times \\ 
Scheme(4) \times Clipping(2) \\ 
\times Granularity(2) \times Mixed\;Precision(2)       
\label{eq:overall_quantization}
\end{multline}
The time cost of the accuracy measurement for a configuration exploration ranges from minutes to hours, depending on hardware platforms. 
\news{As listed in Table~\ref{tab:measurement_time}, we quantified time costs of measuring accuracy depending on the processors capability of the target devices.
Hence, the accuracy measurement across the six models took 0.12-0.58 h on GPU(2080ti), 0.51-12.05 h on CPU(i7-8700), and 10.54-374.15 h on CPU(a53).}
Even though we restrict the possible search space by applying a configuration to the whole layers, an exhaustive search that covers all the combinations of the configurations may take few days when it is performed on the edge and mobile systems. In the following subsection, we elaborate each quantization configuration.

\begin{table}[]
	\centering
	\caption{\news{Time cost for measuring Top1 accuracy depending on target devices.}}
\begin{tabular}{cccc}
	\toprule
	\multirow{2}{*}{\textbf{Model}} & \multicolumn{3}{c}{\textbf{Measurement time (hours)}}\tabularnewline
	\cline{2-4} \cline{3-4} \cline{4-4} 
	& \textbf{CPU(a53)} & \textbf{CPU(i7-8700)} & \textbf{GPU(2080ti)} \tabularnewline
	\hline 
	\textbf{MN} & 26.14 & 1.84 & 0.22\tabularnewline
	\textbf{SHN} & 10.53 & 0.51 & - \tabularnewline
	\textbf{SQN} & 11.56 & 0.52 & 0.03 \tabularnewline
	\textbf{GN} & 374.15 & 12.05 & 0.58 \tabularnewline
	\textbf{RN18} & 53.31 & 2.34 & 0.12 \tabularnewline
	\textbf{RN50} & 126.65 & 4.87 & 0.22 \tabularnewline
	\bottomrule
\end{tabular}
	\label{tab:measurement_time}
\end{table}

\subsection{\news{Calibration Cache}}
\news{There are many ways to generate a calibration cache. The generated calibration cache depends on the given images during the calibration phase, as shown in Fig.~\ref{fig:overview}. Moreover, the considered images for the calibration phase are selected by the \textit{image selector}. To shirk the search space, the image selector randomly chooses images from the ImageNet training dataset using three parameters: 1, 1,000, 10,000. Therefore, three calibration caches were used.
}

\subsection{Scheme}
Considering an efficient code generation on multiple hardware platforms, we focus on a uniform integer quantization. Regarding uniform quantization, the schemes to map the real values to the integers are composed of four linear methods: \textit{asymmetric}, \textit{symmetric}, \textit{symmetric with uint8}, and \textit{symmetric power2}.

\textbf{Asymmetric}: This scheme stands for affine. The float range is converted to [$-2^{n-1}$,$2^{n-1}-1$], where $qmin$ is -128 and $qmax$ is 127. 
This scheme fully uses the presentation capability of \textit{int8}. 
The quantizer of this scheme is defined by the following:
\begin{align}
\begin{split}
x _{i8} & = Quant(x _{fp32} ) = ROUND \left(\frac{x_{fp32}}{scale}+zero\;point \right),
\end{split}
\label{lab:asymmetric}
\end{align}
\news{where $x _{fp32}$ is a real value (fp32) and $x _{i8}$ is 8 bit-width of a signed integer value.
In equation~(\ref{lab:asymmetric}), $scale$ is defined as}

\begin{align}
scale &= \frac{max_{fp32}-min_{fp32}}{2^{n}-1},
\end{align}
\news{where $max_{fp32}$ and $min_{fp32}$ are maximum and minimum values in the current feature map, respectively.
In equation~(\ref{lab:asymmetric}), $zero\;point$ is defined as:}
\begin{align}
zero\;point & = -ROUND(\frac{min_{fp32}}{scale}) - 2^{n-1}
\end{align}
\news{The dequantizer of this scheme is defined as follows:}
\begin{align}
\begin{split} 
x_{fp32} & = Dequant(x _{i8}) = scale \cdot x _{i8} - zero\;point
\end{split}
\end{align}

\textbf{Symmetric}: The real zero directly maps to the quantized zero. 
It does not convert the $min$ and $max$ of the \textit{fp32} range to the quantized range (\textit{int8}).
Instead, the absolute maximum value between the min and max of the \textit{fp32} range is used to set the $qmin$ and $qmax$. 
Symmetric scheme is more efficient than that of the asymmetric because it does not use $zero\;point$. 
However, the symmetric scheme can result in a severe quantization error when the minimum and maximum values are significantly different.
\news{The quantizer of the symmetric scheme is defined as}
\begin{align}
x _{i8} &= Quant(x _{fp32} )=ROUND (\frac{x _{fp32}}{scale})
\label{lab:symmetric}
\end{align}
\news{In equation~(\ref{lab:symmetric}), $scale$ is defined as}
\begin{align}
scale &= \frac{MAX(ABS(x _{fp32}))}{2 ^{(n-1)} -1},
\end{align}
\news{where $MAX(ABS(x _{fp32}))$ is an absolute maximum value among real values in the current feature map.
The dequantizer of this scheme is defined as follows:}
\begin{align}
x_{fp32} & = Dequant(x _{i8} ) = scale \cdot x _{i8}
\end{align}

\textbf{Symmetric with uint8}:
\news{
This scheme is a combination of asymmetric and symmetric schemes. This scheme adaptively switches the quantization method depending on the distribution of real values.
By configuring the offset depending on whether negative values exist, this scheme achieves a computation overhead that is less than or equal to that of asymmetric scheme, and an accuracy higher than or equal to that of symmetric scheme.}
In this scheme, $zero\;point$ is determined in either zero or $-128$. 
\news{The determined $zero\;point$ enables the quantization scheme to be either symmetric($zero\;point=0$) or asymmetric ($zero\;point=-128$). 
In the code level, \textit{uint8} ranges are represented by combining \textit{int8} ranges and the $zero\;point$ of -128. 
The quantizer of this scheme  is defined as follows:}
\begin{align}
x _{i8} &= Quant(x _{fp32} )=ROUND (\frac{x _{fp32}}{scale}+zero\;point)
\label{lab:symmetric_uint8}
\end{align}

\news{In equation~(\ref{lab:symmetric_uint8}), $scale$ is defined as}
\begin{align}
scale &= \frac{MAX(ABS(x _{fp32}))}{2 ^{n} -1},
\end{align}
\news{where $MAX(ABS(x _{fp32}))$ is an absolute maximum value among real values in the current feature map.
In equation~(\ref{lab:symmetric_uint8}), $zero\;point$ is defined as}
\begin{align}
zero\;point &=
\begin{cases}
    -128, & \text{if $min_{fp32}>=0$}\\
    0, & \text{otherwise}   
\end{cases} 
\end{align}
\news{The dequantizer of this scheme is defined as follows:}
\begin{align}
x_{fp32} & = Dequant(x _{i8} ) = scale \cdot x _{i8}-zero\;point
\end{align}

\textbf{Symmetric with power of two-scale}: This is similar to the symmetric. 
This scheme represents the quantized ranges by mapping real zero to the quantized zero. 
In addition to the zero mapping, this scheme converts a real scale to approximately a power of two-scale. 
By doing that, the power of two-scale quantizer substitutes multiplications by bit-shift operations. 
The multiplication elimination makes the hardware design simpler and leads to a better performance although it results in a poor representation of the quantized range.
\news{In addition, only integer operators are used in the entire inference. For that reason, the quantized model with \textit{symmetric with power two-scale} is deployed on integer-only hardware.}
The quantization operation is defined by the following:
\begin{equation} 
scale=2^{\ceil*{\log_2{\frac{MAX(ABS(x _{fp32}))}{2^{(n-1)} -1}}}}
\end{equation}

\news{In summary, the four schemes constitute a trade-off between inference latency and quantization error. As listed in Table~\ref{table:schemes}, we have classified the advantages and drawbacks into four metrics based on the following questions: (i) How precise is the quantization mapping? (fine-grained mapping); (ii) How well is the skewed distribution handled (robustness to skewness); (iii) How much is the execution overhead of quantization? (low computation); and (iv) Can the quantization computation be solely represented with integer operators? (integer-only hardware). 
The comparison reveals that the four methods complement each other because each scheme has its advantages and disadvantages; thus, there is no superior among these four schemes.
}

\begin{table}[t]
\centering
\caption{\news{Comparison of quantization schemes. Three symbols \cmark, \pmark~and, \xmark~denote full, partial, and no supports respectively.}}
\label{table:schemes}
\resizebox{\columnwidth}{!}{
\begin{tabular}{ccccc
}
\toprule
\multicolumn{1}{c}{\textbf{Schemes}}  &
\multicolumn{1}{c}{\textbf{Fine-grained mapping}} & \multicolumn{1}{c}{\textbf{Robustness of skewness}} & \multicolumn{1}{c}{\textbf{Low computation}} & \multicolumn{1}{c}{\textbf{Integer-only HW}} 
\\   \midrule
Asymmetric & \cmark & \cmark & \xmark & \xmark \\ 
Symmetric & \pmark & \xmark & \cmark & \xmark \\
Symmetric w/ uint8 & \pmark & \pmark & \pmark & \xmark \\
Power of two-scale & \xmark & \xmark & \cmark & \cmark \\ \bottomrule
\end{tabular}
}
\end{table}

\subsection{Clipping}
Without the retraining step, a uniform quantization causes an accuracy drop. 
Accuracy loss mainly stems from the Gaussian shape of the distributions for weights and activations of the pre-trained neural networks~\cite{lin2016fixed,han2015deep}.
Considering such a characteristic of the distributions, a few weights and activations are sparsely spread as outliers. The outliers in a long tail make a uniform quantizer assign few quantization levels to small values and too many to large ones. This skewness of the distributions leads to significant accuracy degradation~\cite{zhao2019improving,kris_whitepaper2018}.

To rectify this problem, the Glow compiler allows clipping the range of the distributions for weights and activations. This method chooses a clip threshold which (approximately) minimizes the Kullback–Leibler (KL) divergence between the floating-point and quantized~\cite{migacz20178}.

\subsection{Granularity}
Considering the quantization, we decided on how far the scale value should be shared among the tensors. We refer to this choice as quantization granularity. We consider two kinds of granularity for the quantization: tensor-wise and channel-wise. Granularity shows the trade-off between accuracy and latency. Fine-grained granularity requires more computation because of the increasing multiplications~\cite{wu2020integer}.
Furthermore, a convolution containing a wide range of weight values should consider the channel-wise to compute the scale for quantization. Therefore, granularity is determined by examining the granularity impact on accuracy and latency.

\subsection{Mixed-precision}
As a part of the extended Glow, we implemented layer-wise mixed precision by extending the Glow compiler. The original Glow compiler does not support the mixed-precision at the layer level. Instead of considering all the layers for the mixed-precision, we only keep the first and last layers of the original precision (\textit{fp32}). 
This is because an experimental result of a previous study~\cite{wu2020integer} demonstrates that the first and last layers are the most sensitive to the quantization error.

\section{Modeling: Parameter Search using eXtreme Gradient Boosting}\label{sec:modeling}

\subsection{Problem Definition}
The quantization configuration search is performed using historical data previously found in other CNN models regardless of having to search for a new quantization configuration from a random initial point. 
We hypothesize that the quantization configurations of a CNN model is related to other configurations. 
Considering the model, a next configuration can be predicted and the result sends feedback to the online training process to update the model.

As described in equation (\ref{eq:tuning}), our tuning problem can be formulated, considering two kinds of features as demonstrated below:
\begin{align}
s_{opt}^{*} &= \argmax_{s \in S_{e}} f(g(e,s))
\label{eq:tuning}
\end{align}
First is the block expression of the CNN as denoted by $e$. 
The CNN block expression $e$ consists of the following operations: the number of layers, convolutions, activation functions, skip-layers, and depth-wise and pointwise convolutions. 
These kinds of blocks are based on the predefined common structures in the neural architecture search~\cite{wu2019fbnet,wan2020fbnetv2}.
The blocks have been used to reduce the search time and to find a good model in the neural architecture search~\cite{wu2019fbnet,wan2020fbnetv2}.

We generate multiple quantized models that have different accuracies for a given $e \in \mathcal{E}$.
We use $S_{e}$ to denote the space of quantization configurations from $e$ to the quantized models. 
For instance, if $s \in S_{e}$ let $x = g(e, s)$ is the generated quantized model, $g$ represents the \textit{Glow extension} that generates the quantized tensor IR from $e,s$. 
We aim to maximize $f(x)$, which is the accuracy of the quantized models on the target hardware.
Particularly, we verified if an output for $f(x)$ could measure accuracy by running experiments on the hardware. 
For a given $g,e,S_{e},f$, the accuracy of the quantized models are demonstrated using equation (\ref{eq:tuning}).

\begin{figure}[t]
	\centering
	\includegraphics[width=1\columnwidth]{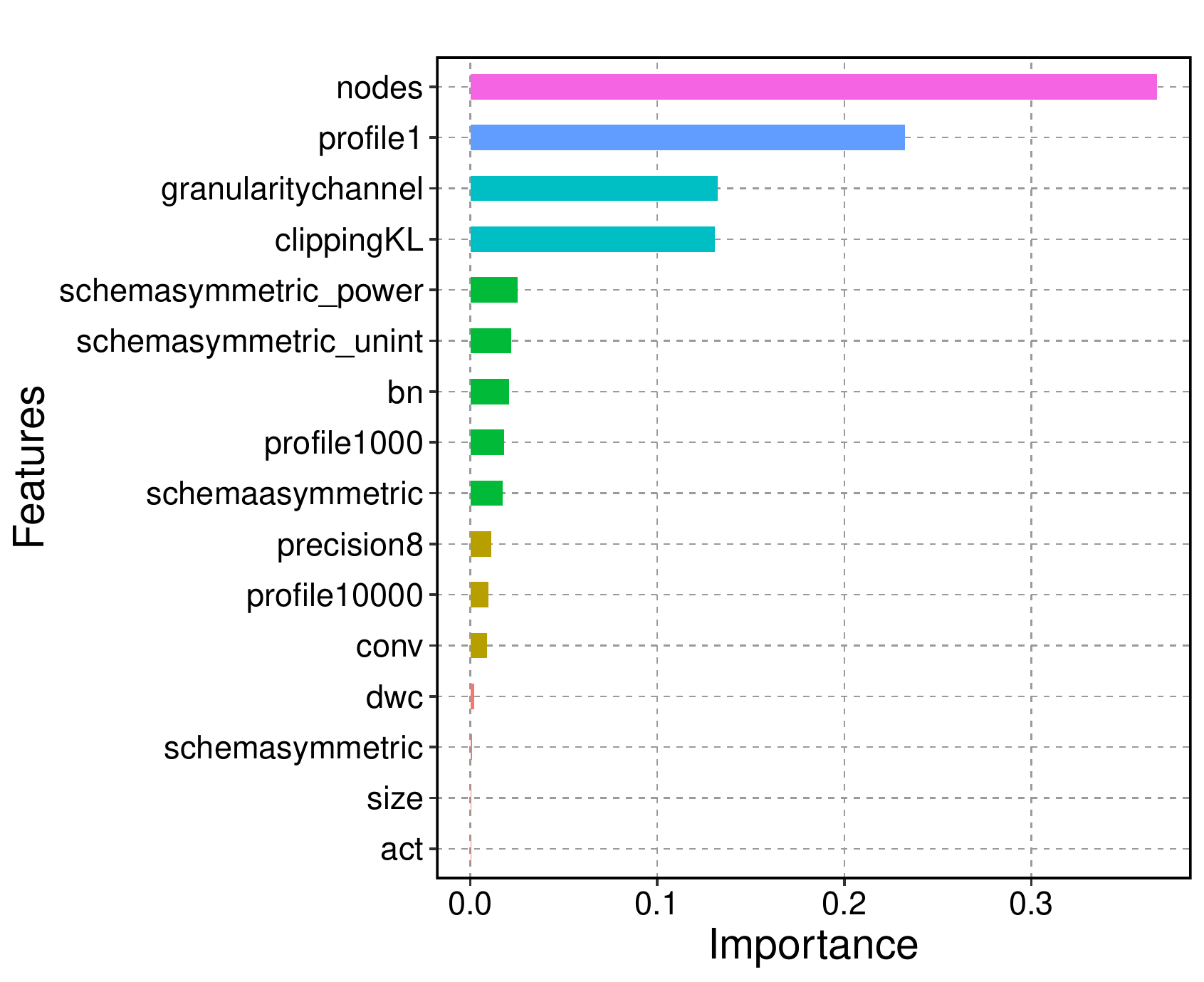}
	\caption{Ranking of the features used for the modeling.}
	\label{fig:importance}
\end{figure}

\subsection{Auto-tuning Algorithm}

\begin{figure*}[t]
	\centering
	\includegraphics[width=2\columnwidth]{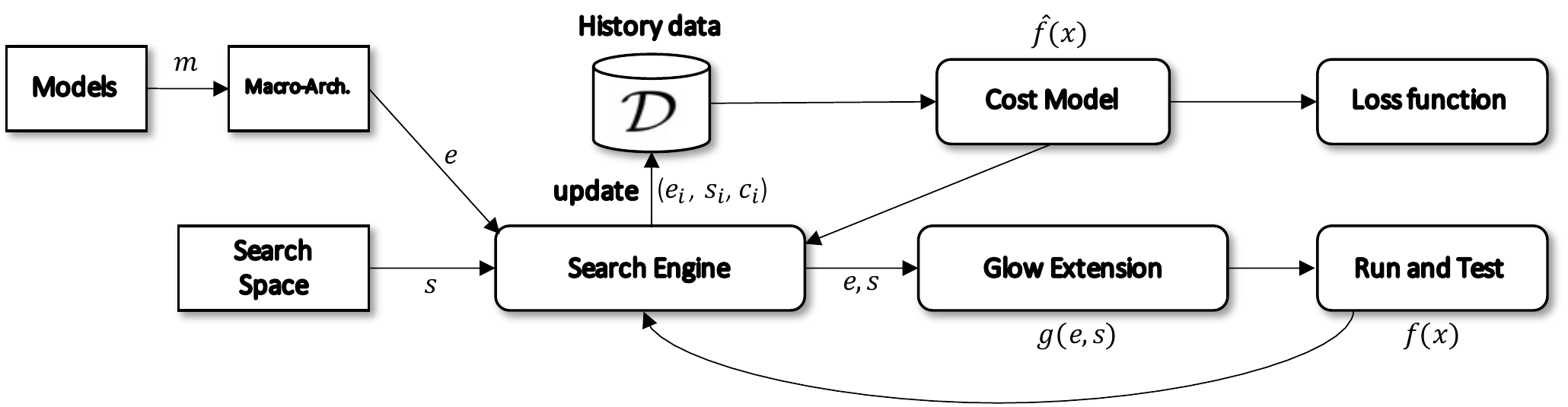}
	\caption{Auto-tuner to efficiently explore quantization configurations.}
	\label{fig:costmodel}
\end{figure*}

We propose a machine learning (ML)-based auto-tuning to search the quantization configurations. 
Fig.~\ref{fig:costmodel} shows the components of the auto-tuning and how they interact with one another. 
To predict the accuracy of quantized model $x$, the cost model $\hat{f}(x)$ is trained using the historical data. 
The search engine generates a new quantized model by using the Glow extension and measures its accuracy on the hardware. 
The accuracy of the quantized model is saved in a database as demonstrated below: $\mathcal{D}=\{(e_{i},s_{i},c_{i})\}$.
\news{As shown in Fig.~\ref{fig:overview}, $e_{i},s_{i},$ and $c_{i}$ denote the measured accuracy, the extracted model architecture, and the explored configuration, respectively.}
The collected data could be used to train $\hat{f}$. 
The following subsections elaborates objective function to train a statistical cost model and the design choices of each component.

\subsubsection{Cost Model and Objective Function}
\news{XGBoost is an extension and improvement of the gradient tree boosting (GBT) algorithm. 
The characteristic of XGBoost is scalability, efficiency, sparsity-aware fitting, and well-supported libraries~\cite{chen2016xgboost}.
To this end, the XGBoost is widely adopted to solve real-world problems such as security~\cite{dhaliwal2018effective,chen2018xgboost}, fault detection~\cite{zhang2018data}, drug discovery~\cite{ji2019five}, and disease diagnosis~\cite{ogunleye2019xgboost,budholiya2020optimized}.
In addition to academic fields, XGBoost is the winning algorithm in Kaggle challenges~\cite{sagi2018ensemble}.}

\news{We selected the XGBoost algorithm to train the data successfully found in the other CNN models and to predict the configuration arising from the most accurate model generation in the next quantization step.
With the cost model $\hat{f}\left(x\right)$ of XGBoost, we could estimate the accuracy of each quantized model x. Specifically, the cost model $\hat{f}\left(x\right)$ of XGBoost represents a tree ensemble model that uses K additive functions defined as}
\begin{equation}
    \news{\hat{f}(x_i) = \hat{y_i} = \sum_{k=1}^{K}{f_{k}(x_i)} , f_k \in F}
    \label{eq:xgboost_cost}
\end{equation}
\news{For a given quantized model $x_i$, $\hat{f}(x_i)$ is the predicted accuracy $\hat{y_i}$. 
The value of the $i$-th instance is $f_{k}(x_i)$ at the $k$-th tree. Moreover, the values of the space of trees are denoted by the function $F$.}

\news{To train the cost model $\hat{f}(x)$, we follow the regularized objective function and the optimization method introduced in the original paper~\cite{chen2016xgboost}.
The regularized objective function of XGBoost is defined as}
\begin{equation}
    \news{Obj = \displaystyle\sum_{i=1}^{N}{\mathcal{L}(\hat{y_i},y_i)} + \displaystyle\sum_{k=1}^{K} \Omega(f_k)}
    \label{eq:xgboost_objective}
\end{equation}
\news{Equation~(\ref{eq:xgboost_objective}) consists of two parts: the loss function ($\mathcal{L}(\hat{y_i},y_i)$) and the regularization function ($\Omega(f_k)$).
$\mathcal{L}(\hat{y_i},y_i)$ is a differentiable convex function that computes the difference between the prediction $\hat{y_{i}}$ and the true label $y_i$.
The differentiable convex functions are mean square error and Logistic loss.
The regularization function $\Omega(f_k)$ for the $k$-th tree is then defined as}
\begin{equation}
    \news{\Omega(f_{k})=\gamma T + \frac{1}{2}\lambda ||w||^2,}
    \label{eq:xgboost_regularization}
\end{equation}
\news{where $T$ is the number of leaves in a tree, $\gamma$ lies between 0 and 1 and is multiplied by $T$ to reduce the complexity of each leaf, and $\lambda$ is a parameter that scales the penalty to avoid the overfitting.}

\news{The regularized objective function in equation~(\ref{eq:xgboost_objective}) cannot be optimized using traditional methods.
Instead, the cost model is trained using an additive method. 
At the $t$-th step, the $\hat{y_{i}}^{(t)}$ is optimized as follows:}
\begin{equation}
    \news{Obj^{{(t)}} = \sum_{i=1}^{n}{\mathcal{L}\left( y_i,\hat{y_i}^{(t-1)} + f_{t}(x_i)\right) + \Omega(f_t)}}
    \label{eq:xgboost_objective_detail}
\end{equation}
\news{Each $f_{t}$ denotes to an independent tree generated by instance $i$ in the $t$-th step.
The additive method combines all $f_{t}$ that maximally improves the objective function in equation~(\ref{eq:xgboost_objective}).
To reduce the computation cost of the objective function,  equation~(\ref{eq:xgboost_objective_detail}) is transformed using the second-order Taylor approximation as follows}
\begin{equation}
\news{    Obj^{{(t)}} \simeq \sum_{i=1}^{n}{\left[\mathcal{L}(y_i,\hat{y_i}^{(t-1)}) 
    + g_{i}f_{t}(x_{i}) 
    + \frac{1}{2}h_{i}f_{t}^{2}(x_{i})\right]
    +\Omega(f_t)},}
    \label{eq:xgboost_taylor}
\end{equation}
\news{where $g_i$ and $h_i$ are defined as}
\begin{equation}
\begin{split}
\news{g_i = \partial_{\hat{y}_{i}^{(t-1)}}\mathcal{L}(y_i,\hat{y_i}^{(t-1)})} \\
\news{h_i = \partial_{\hat{y}_{i}^{(t-1)}}^2\mathcal{L}(y_i,\hat{y_i}^{(t-1)})}
\end{split}
\label{eq:xgboost_partial}
\end{equation}
\news{We could simplify equation~(\ref{eq:xgboost_partial}) by removing the constant terms. 
At step $t$, the final objective function depends on the first and second-order gradients and is defined as}
\begin{equation}
    \news{\tilde{Obj^{{(t)}}} = \sum_{i=1}^{n}{\left[g_{i}f_{t}(x_{i}) 
    + \frac{1}{2}h_{i}f_{t}^{2}(x_{i})\right]
    +\Omega(f_t)}}
    \label{eq:xgboost_simplifiedobject}
\end{equation}

\news{Finally, with the simplified objective function in equation~(\ref{eq:xgboost_simplifiedobject}), we could iteratively evaluate the model performance after a certain node split in a tree.
If the tree model performance is improved after splitting, this change will be accepted; otherwise, the split will be stopped.
In this manner, the optimal splitting point for each tree to minimize the objective function is determined; hence, the regularization term remedies overfitting during training.
This is the working principle of XGBoost.}

\subsubsection{Training Objective Function with hyperparameters} 
The XGBoost models have hyper-parameters which can be set in order to customize the model for a specific dataset. 
To train the XGBoost model, we consider several factors including the hyper-parameters, preprocessing, and loss functions.

First, the hyper-parameters were \textit{Eta} and \textit{gamma}. 
To feed the dataset to the XGBoost model, we considered the preprocessing of two kinds of features: a model-arch ($e_{i}$) and a configuration ($s_{i}$). 
Second, the preprocessing can be determined using categorical or one-hot encoding. 
In this study, we consider one-hot encoding features for the preprocessing because it shows better accuracy than the categorical ones. 
Third, the possible loss functions for the training are \textit{rank} and \textit{regression}. 
To apply the \textit{rank} loss function in the training process, it is necessary to add rank the information, and the ranked information can be grouped by a type of the CNN models or whole data. The result of the comparison between the two loss functions shows that regression achieves a better search result with a lower number of trials. Therefore, we consider regression function.

To understand the value of the selected features in the construction of the XGBoost model, we performed the analysis of the feature importance using the XGBoost library in \textit{R}. 
The significant score of the features is simply calculated using purity (the Gini index)~\cite{hastie2009elements}. 
Fig.~\ref{fig:importance} shows the result of the importance of the feature.
As a result, the number of nodes, calibration (profile), granularity, and clipping is important, considering the predicted accuracy of the XGBoost model.

\begin{algorithm}[t]
  \caption{\textsf{Quantune}: Search for the Optimal Configuration.}\label{xgboost}
  \hspace*{\algorithmicindent}\textbf{Input:} Macro-arch. blocks $e$ \\
  \hspace*{\algorithmicindent}\textbf{Input:} Search space $S_e$ \\
  \hspace*{\algorithmicindent}\textbf{Output:} $s^{*}_{opt}$ 
  
  \begin{algorithmic}[1]
  \LState $\mathcal{D} \gets \emptyset$ \Comment{The collected data will be contained}
  \While{$n\_trials<max\_n\_trials$}
    \LState $s \gets$ A top candidate in unexplored $S_{e}$ using $\hat{f}$
    \LState $c\gets f(g(e,s))$
    \LState $\mathcal{D}\gets \mathcal{D} \cup \{(e,s,c)\}$
    \LState update $\hat{f}$ using $\mathcal{D}$
    \LState $n\_trials \gets n\_trials + 1$ 
  \EndWhile
  \LState \Return $s^{*}_{opt} \gets$ history best quantization config.
  \end{algorithmic}
\end{algorithm}

\subsubsection{Search Engine}
The search engine seeks the optimal configuration of the quantization as described in Algorithm~\ref{xgboost}. 
Algorithm~\ref{xgboost} takes macro-arch blocks $e$ and search space $S_{e}$ as its inputs. 
It produces the optimal configuration for quantization as an output. 
At each iteration, the engine picks a candidate based on $\hat{f(x)}$ and queries $f(x)$ on the accuracy of quantized model. 
We enumerate the entire space of $S_{e}$ and pick the top candidate. 
The top candidate is not explored in the previous step. 
To accelerate convergence in search step, we apply transfer learning.

The loop iterates over possible optimal configurations. 
We set the maximum iteration $max\_n\_trials$ = search space as described in equation (\ref{eq:overall_quantization}).
In Algorithm~\ref{xgboost}, lines 2–8 describe the steps needed to obtain an optimal configuration from the possible configuration space $S_{e}$. At each iteration, the following steps take place sequentially. 
First, a top candidate is selected using $\hat{f}$, considering diversity. 
Therefore, we select a candidate from unexplored configurations and try to measure the accuracy of the quantized model on the real target hardware.

\section{Evaluation}
We conducted three experiments: 1) the variation of accuracy depending on the quantization configurations, 2) effectiveness of the XGBoost based parameter search in search time and accurate model generation, and 3) benefits of the inference time in low precision representation.

\subsection{Diversity in quantization configurations}
Considering the broad selection of the quantization configurations, it is not obvious to determine which choice mostly reduces the quantization error. To find out the best configuration, we empirically explored various choices for \textit{int8} quantization of the six models. 
As described in equation (~\ref{eq:overall_quantization}), the search space includes a large number of configurations. 
We experimentally explored all the combinations that map \textit{fp32} to \textit{int8} to show that there was no universal configuration that was always applied to attain the most accurately quantized models. 
Fig.~\ref{fig:combi_qm} shows the Top1 accuracy. 
The relative error which substrates the quantized accuracy from the baseline ranges from -71.72\% to 0.19\% across all the combinations. Considering the exploration results of all the configurations, we found the following insights. The accuracy of the quantized models with various configurations varies. Consequently, there is no clear solution to be applied in all the cases. Whether clipping is applied depends on the amount of calibration data. If calibration is performed with a small number of samples, there are relatively few outliers. In this case, the quantization error can be reduced by using full ranges of tensors without clipping. In contrast, clipping reduces the quantization error by increasing the number of sample images during the calibration.

To determine the diversity among the quantization configuration at the accurately quantized models, the results are selected within relative error of 1\%. The industrial margin for quantization error is 1\% accuracy drop because a de-facto benchmark known as MLPerf~\cite{reddi2020mlperf} allows the quantized model to degrade the accuracy within -1\%.
Therefore, the following analysis for diversity is based on 1\% accuracy drop across all the quantized models.

The Shannon-entropy equation~(\ref{eq:entropy}) is used to show diversity index in each quantization configuration.
\begin{align}
H(X) = H(p) = - \sum_{i}{p(x_i) \times \log{p(x_i)} }
\label{eq:entropy}
\end{align}
The analysis demonstrates the number of configurations that generate the quantized models that meet the industrial quality and diversity index. Each column in Table~\ref{table:entropy} shows the diversity index that represents the diversity of the configurations. 
The diversity indexes for all models within 1\% range from 0.50 to 1.80 across the calibration, scheme, clipping, granularity, and mixed-precision.
If the entropy is zero, there is no uncertainty. This indicates that there is no obvious configuration to generate the optimal quantized models because all the configurations are not zero entropy. 
Therefore, it is hard to intuitively and manually select a configuration for the quantization.

\begin{table}[t]
\centering
\caption{Diversity analysis of quantization configurations. For the analysis, the quantization configurations that achieve accuracy loss within 1\% are used.}
\label{table:entropy}
\resizebox{\columnwidth}{!}{
\begin{tabular}{rrrrrr
}
\toprule
\multicolumn{1}{c}{\textbf{Precision}}  &
\multicolumn{1}{c}{\textbf{Calibration}} & \multicolumn{1}{c}{\textbf{Granularity}} & \multicolumn{1}{c}{\textbf{Clipping}} & \multicolumn{1}{c}{\textbf{Scheme}} & \multicolumn{1}{c}{\textbf{\# of Samples}} 
\\   \midrule
0.50 & 1.43 & 0.99 & 0.98 & 1.80 & 71 \\ \bottomrule
\end{tabular}
}
\end{table}

\begin{figure*}[t]
	\centering
    \includegraphics[width=2\columnwidth]{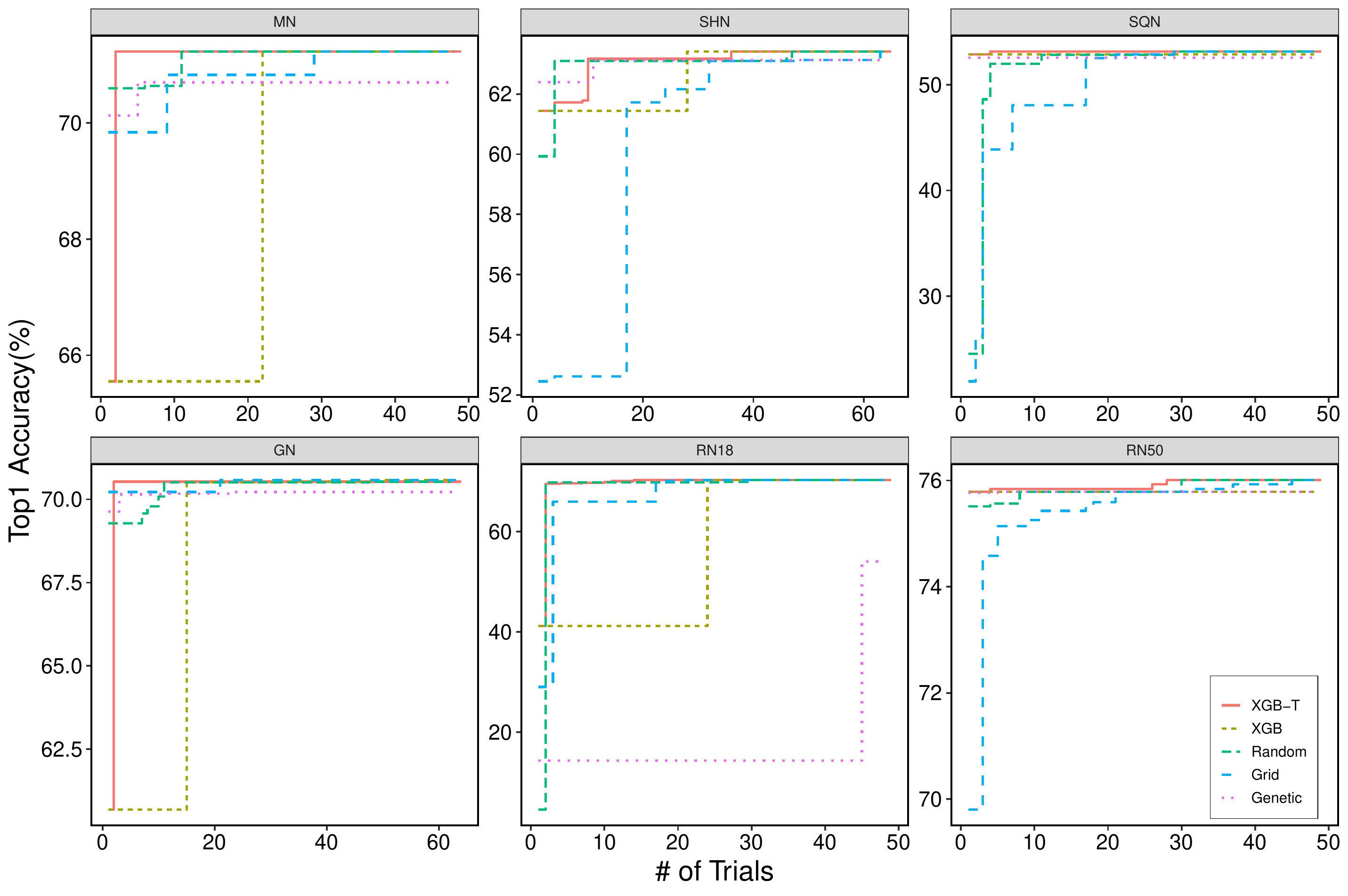}
	\caption{\news{Comparison of different search algorithms, considering the convergence speed and Top1 accuracy on the six CNN models.}}
	\label{fig:trials_overall}
\end{figure*}

\begin{figure*}[t]
	\centering
	\includegraphics[width=1.8\columnwidth]{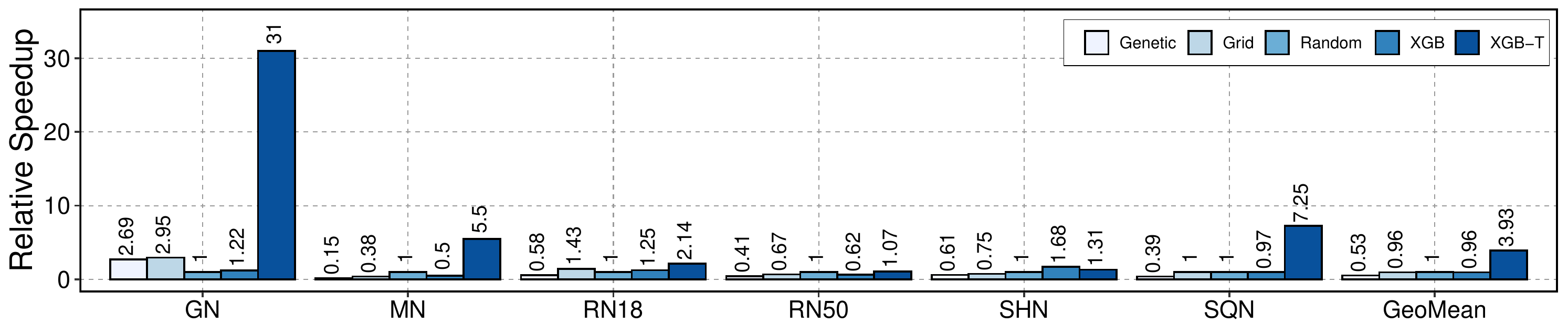}
	\caption{\news{Relative speedups of convergence over the random search.}}
	\label{fig:relative_trials}
\end{figure*}

\begin{figure}[t]
	\centering
	\includegraphics[width=1\columnwidth]{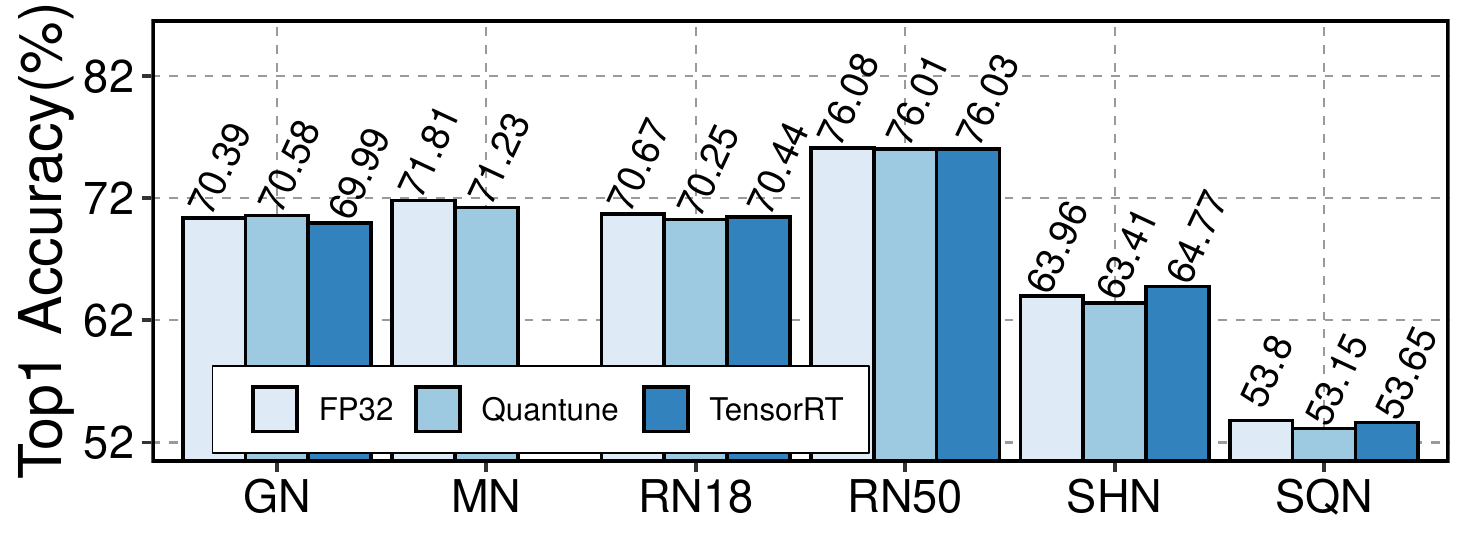}
	\caption{\textsf{Quantune} vs. TensorRT - \textsf{Quantune} achieves a comparable accuracy of the quantized CNN models against the off-the-shelf compiler (TensorRT) on the NVIDIA GPU.}
	\label{fig:trtacc}
\end{figure}

\begin{figure}[t]
	\centering
	\includegraphics[width=1\columnwidth]{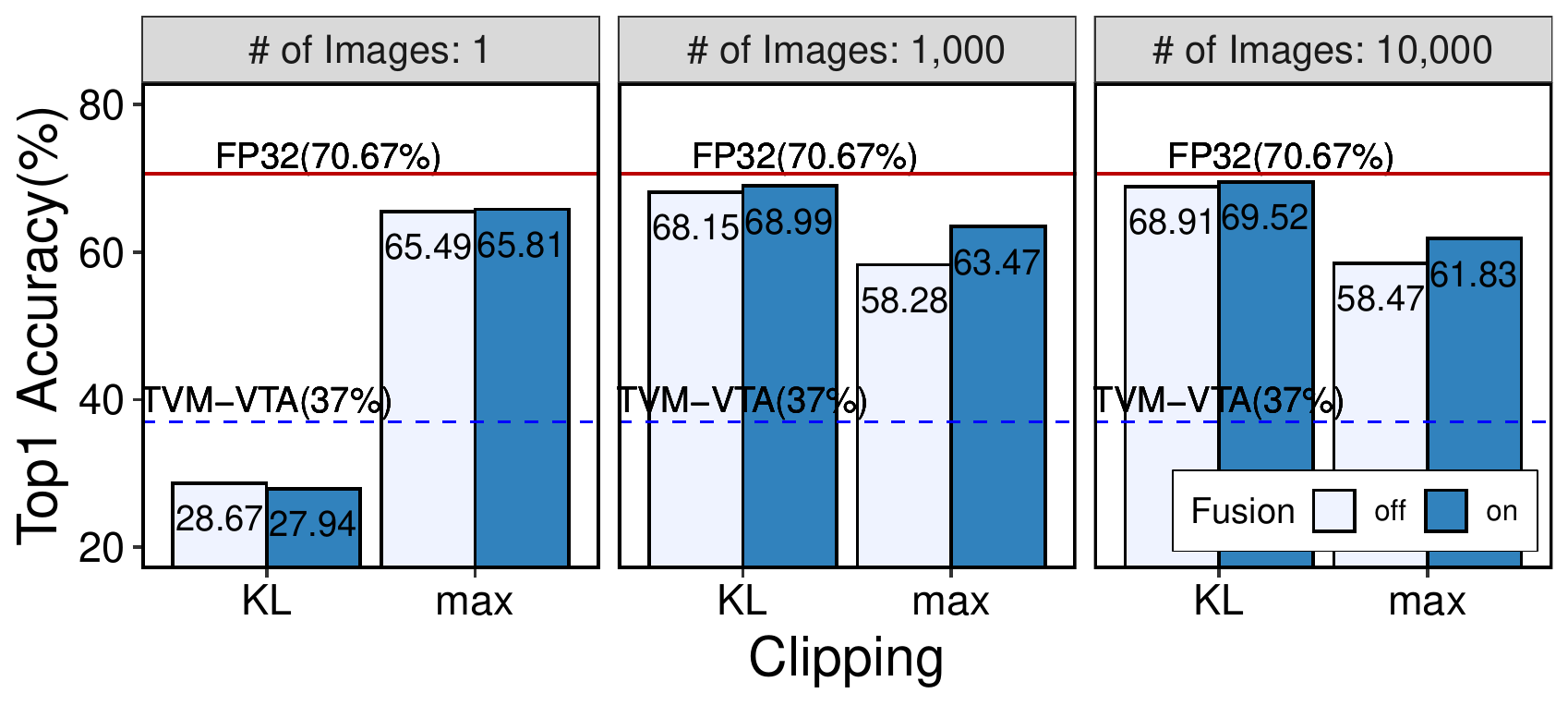}
	\caption{\textsf{Quantune} vs. TVM-VTA - \textsf{Quantune} leads to the significant improvement in accuracy by approximately 32.52\% as against the TVM-VTA~\cite{moreau2018leveraging}. The solid-line indicates full-precision accuracy (\textit{fp32}). The dashed-line indicates the TVM-VTA accuracy.}
	\label{fig:vtaacc}
\end{figure}

\begin{figure}[t]
	\centering
	\subfloat[][ARM A53]{\label{lat:lat_a53}\includegraphics[width=.45\textwidth]{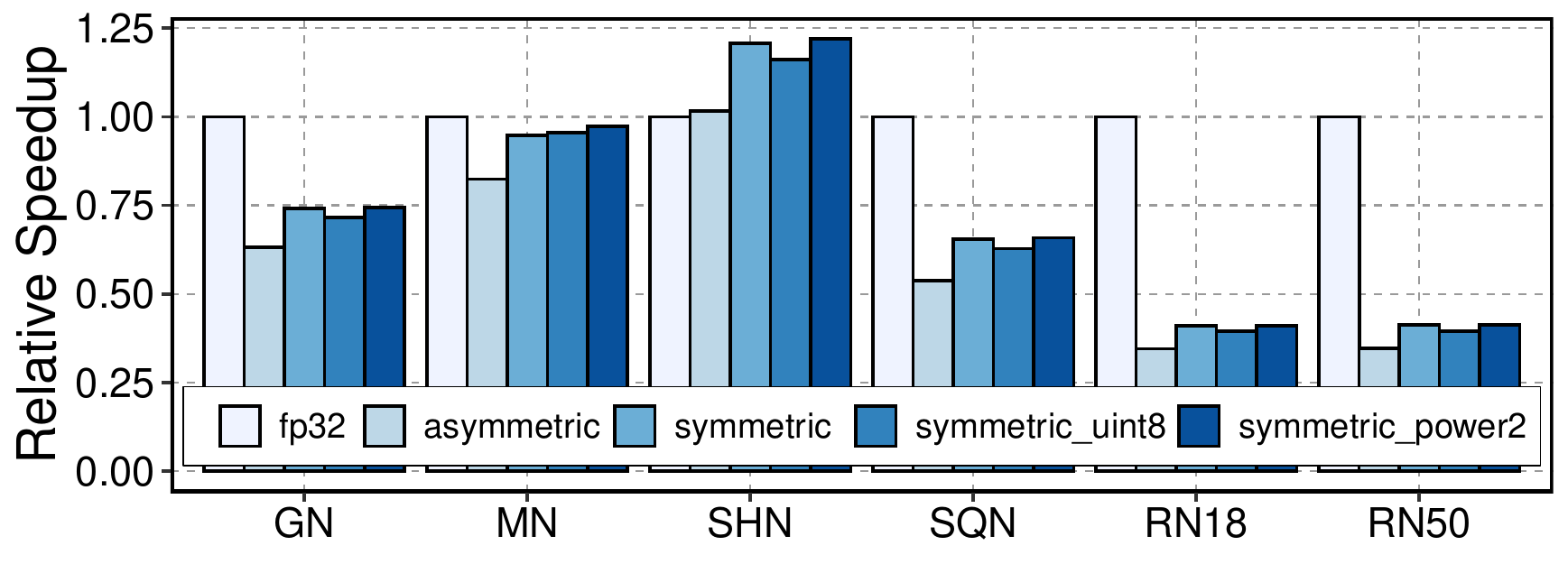}}\vfill
	\subfloat[][NVIDIA 2080ti ]{\label{lat:lat_gpu2080ti}\includegraphics[width=.45\textwidth]{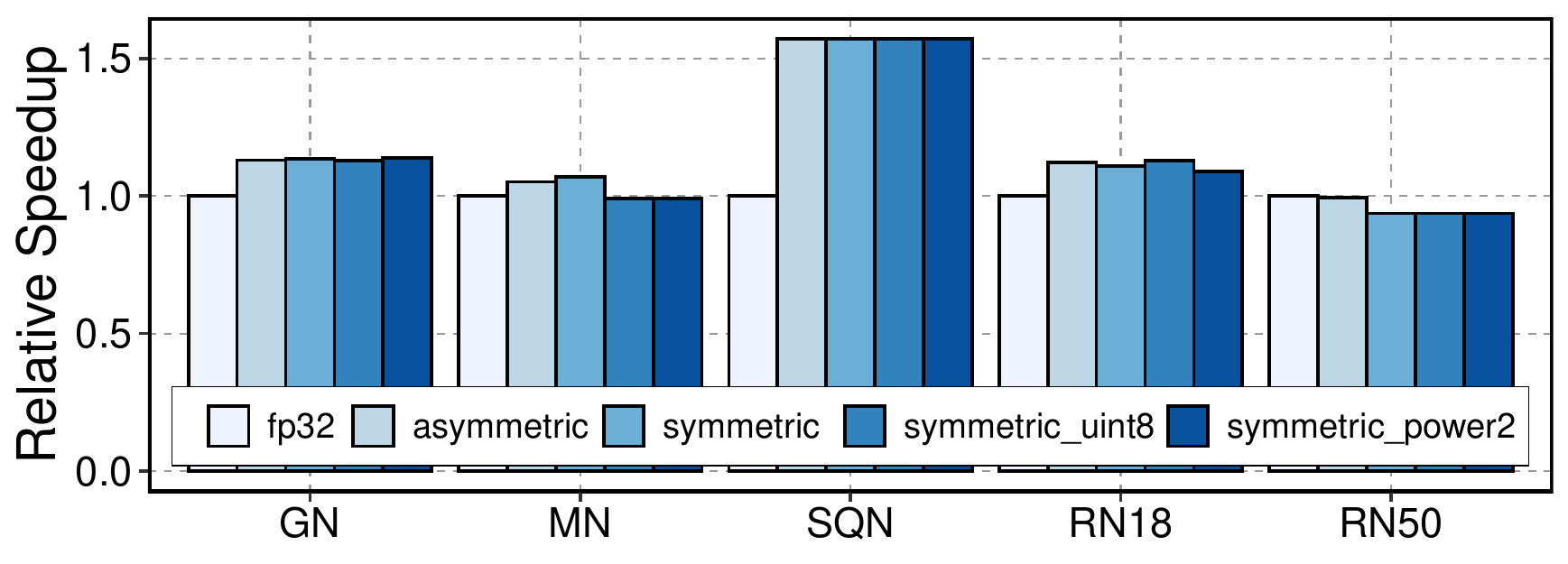}} \vfill
	\subfloat[][Intel i7-8700]{\label{lat:lat_i78700}\includegraphics[width=.45\textwidth]{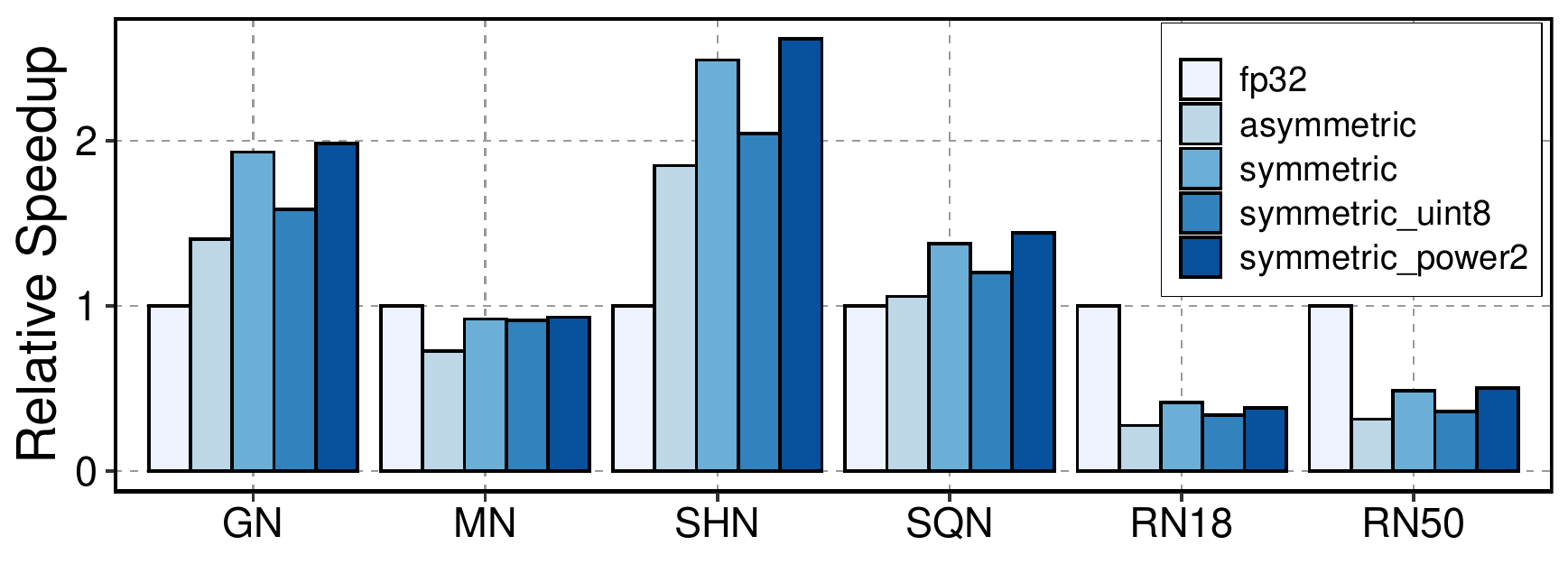}}
	\caption{Comparison of relative speedup between the original and quantized models on ARM A53, NVIDIA 2080ti, and Intel i7-8700.}
	\label{fig:latency}
\end{figure}

\subsection{Efficiency of XGBoost based configuration search}
From the experiment, the quantization configuration for the most accurately quantized models varies, depending on the type of models. However, the search space for the configuration selection is large as shown in equation~(\ref{eq:overall_quantization}). 
To reduce the search time, we devise a \textsf{Quantune} that seeks the optimal configuration for the generation of the quantization model based on the XGBoost.

In this section, we show how many trials are required to obtain the optimal quantized model with the optimal configuration. 
To show the efficiency of the \textsf{Quantune}, we compared it to \news{four} algorithms.
The \textit{random} search defines a search space as a combination of hyper-parameter values and randomly selects a point in the range.
The \textit{grid} search specifies a search space as a grid of the hyper-parameter values and samples of a point in the grid. 
\news{The \textit{genetic} algorithm (GA) is an optimization method that exploits the idea of evolution by natural selection.
GA is based on the hypothesis that a new population will be better than the previous one.
To apply genetic algorithm in the quantization configuration search, we implemented the details using the GA package~\footnote{https://cran.r-project.org/web/packages/GA/GA.pdf}.
With the GA package, we defined Top1 accuracy evaluation as the fitness function.
Furthermore, we exploited binary encoding for crossover and mutation.
In other cases, we used the default settings from the GA package.}
We apply the XGBoost in two cases: individual learning and transfer learning.
The individual XGBoost model is initialized with no historical data that are produced from the other CNN models. 
It shows that the XGBoost online-learning starts from the base of the current CNN model-tuning. Practically, the \textsf{Quantune} can collect historical data ($\mathcal{D}$) from previous exploration. 
We combine the XGBoost and transfer learning to effectively use $\mathcal{D}$.

Fig.~\ref{fig:trials_overall} and~\ref{fig:relative_trials} show \news{five} algorithms that search the quantization configurations of the six CNN models. Our proposed search algorithm (\textit{XGB-T}) is able to guide the configuration search to faster convergence for MN, SHN, SQN, GN, RN18, and RN50 by 5.50$\times$, 1.31$\times$, 7.25$\times$, 31$\times$, 2.14$\times$, and 1.07$\times$ for the \textit{random} search and 14.5$\times$, 1.75$\times$, 7.25$\times$, 10.5$\times$, 1.5$\times$, and 1.6$\times$ for the \textit{grid} search, respectively. \news{In addition, XGB-T yielded a 2.13$\times$ to 36.5$\times$ speedup compared with the \textit{genetic} search.
All the improvements are due to the cost model and transfer learning.
In contrast to XGBoost, the other three algorithms work without cost model.} Furthermore, transfer learning allows speedup and attains a higher accuracy. 
Specifically, transfer learning improves the convergence time of the MN, GN, and RN18 relative to the XGBoost by 11$\times$, 25$\times$, and 1.7$\times$, respectively. 
Regarding the SQN and RN50, the \textit{XGB-T} achieved +0.27\% and +0.22\% higher accuracy, respectively, than the \textit{XGB}.

\subsection{Accuracy}
Through the extended Glow stack, the \textsf{Quantune} supports both of the general-purpose units (CPU or GPU) and the VTA. 
Considering both targets, we show how accurate the \textsf{Quantune} generates the quantized models compared to the other existing tools.

First, in the general-purpose targets, the \textsf{Quantune} is directly compared to NVIDIA TensorRT7.2.2 (released in 19 Dec. 2020) on the server-side GPU. 
The TensorRT\footnote{https://developer.nvidia.com/tensorrt} is a well-supported tool by NVIDIA to speed up an inference of the CNN.
For comparison, we develop a test code that generates the quantized models using TensorRT’s post-training quantization and releases it as an open source\footnote{https://github.com/leejaymin/tensorrt\_quantization\_imagenet}.
As shown in Fig.~\ref{fig:trtacc}, the \textsf{Quantune} achieves 0.59\% higher accuracy on GoogleNet slim v4 than the TensorRT. 
Nevertheless, it attains 0.19-1.36 lower accuracy across the four models than the TensorRT.
Considering that the \textsf{Quantune} is a unified open toolchain to support multiple hardware devices, it shows a competitive accuracy compared to the TensorRT which is used only for the NVIDIA GPUs.

Second, for the VTA, we compare the \textsf{Quantune} to the TVM (released in Jul. 2017) on the VTA. Previous results from the VTA-TVM~\cite{moreau2018leveraging} reported a significant accuracy drop (-33.76\%). 
This drastic accuracy drop in the TVM-VTA stems from the choice of a quantization scale for the whole network because a scale value can be imprecise for small values and truncate large values. 
Contrary to the VTA-TVM~\cite{moreau2018leveraging}, the \textsf{Quantune} selects different scales depending on each layer and traverses all the possible configurations.
The search space is different because of the limitations of the VTA hardware. 
In the VTA, the scheme and granularity only support the power of the two-scale and tensor level.
Owing to the architecture support, the fused operator for the convolution and ReLU is executed in consecutive cycles without extra off-chip memory access. 
There are 12 distinct configurations derived from the following possible combinations.
\begin{multline}
Search\;Space(12)=\news{Calibration\; Caches(3)} \times \\ 
Schemes(1) \times Clipping(2) \times \\ 
Granularity(1) \times Fusion(2)       
\label{eq:vta_quantization_space}
\end{multline}
From the results of all configurations, as shown in Fig.~\ref{fig:vtaacc}, \textsf{Quantune} leads to significant improvement in accuracy by approximately 32.52\%. We show that even with a limited scheme and granularity, the best result is 0.73\% lower than the ones on the general-purpose units.

\subsection{\news{Model Size}}
\news{To demonstrate the effectiveness of our model, we measured the size of quantized models depending on the quantization configurations.
Among the considered configurations, \textit{granularity} and \textit{mixed-precision} affect model size.
As listed in Table~\ref{table:model_size}, the size of quantized models varies for four configurations combined with granularity and mixed-precision.
With tensor granularity alone, the compression rate is the highest spanned across six quantized models.
In contrast, the combining channel granularity and mixed-precision shows the lowest compression rate.
The differences between tensor and channel granularity stem from the number of scale factors for quantization. 
Intuitively, channel level quantization requires more scale factors than tensor-level.
In the case of mixed-precision, the model size is dependent on the number of parameters in the first and last layers.}


\begin{table}[t]
\centering
\caption{\news{Comparison of the size of quantized models depending on quantization configuration. Model size means the number of bytes required to save all of the weights in the CNN model.}}
\label{table:model_size}
\resizebox{\columnwidth}{!}{
\begin{tabular}{lrrrrr
} 
\toprule
\multicolumn{1}{c}{\textbf{Model}} & \multicolumn{1}{c}{\textbf{Original}} & \multicolumn{1}{c}{\textbf{Tensor}} & \multicolumn{1}{c}{\textbf{Channel}} & \multicolumn{1}{c}{\textbf{Tensor+Mixed}} & \multicolumn{1}{c}{\textbf{Channel+Mixed}}
\\   \midrule

MN & 13.96MB & 3.54MB & 3.70MB & 7.39MB  & 7.54MB \\
SHN & 5.52MB & 1.42MB & 1.54MB & 3.06MB  & 3.17MB \\ 
SQN & 4.94MB & 1.25MB & 1.29MB  & 1.25MB  & 1.29MB \\ 
GN & 170.60MB & 42.75MB & 43.01MB & 47.36MB & 47.63MB \\ 
RN18 & 47.94MB & 12.91MB & 12.95MB & 14.47MB  & 14.51MB \\ 
RN50 & 102.12MB & 25.61MB & 25.84MB  & 31.79MB  & 32.01MB \\ \bottomrule
\end{tabular}
}
\end{table}

\subsection{Latency}
Our evaluation has focused on the accuracy of the quantized models. 
In this section, we measured end-to-end inference time using the Glow’s code generation (referred to as \textit{CodeGen}) on a server class GPU (an embedded CPU) and a desktop CPU for both the floating point and quantized models. 
As shown in Fig.~\ref{fig:latency}, the inference time of all the quantized models is not improved against the \textit{fp32} models. 
This is because the \textit{CodeGen} in the Glow is not efficiently implemented at the \textit{int8} quantization level for chosen target platforms. 
Previous studies have reported that the naive implemented kernels for quantization can be slower than the original models~\cite{cowan2020automatic,jacob2018quantization,jain2020efficient,jiang2021automated}. 
\news{As mentioned in Section~\ref{table:schemes}, the four schemes are a trade-off between inference latency and quantization error. Therefore, the latency of the quantized models varies for different schemes.}

In the naive implementation, the increasing latency stems from extra operations such as the scales and offsets in each layer of the quantization. In some cases like ShuffleNet, the improvements of latency are attributed to better cache reuse by reducing the weights and activations.

In a nutshell, the quantization method in DL frameworks preserves the accuracy of the quantized models; however, it is insufficient to provide latency improvement on diverse hardware devices. To overcome this challenge, few studies have been proposed~\cite{cowan2020automatic,jacob2018quantization,jain2020efficient,jiang2021automated}. 
An efficient kernel code generation for quantized models is an important research direction that is beyond the scope of our study.

\section{Conclusion}
To enable rapid deployment of quantized models without noticeable accuracy loss, we present \textsf{Quantune}, which can build a model to find optimal configurations for quantization, thereby accelerating the search speed and enhancing the accuracy of the quantized models. The experimental results on real hardware devices show that \textsf{Quantune} achieved 1.3-31 faster convergence time than the three algorithms with 0.07–0.65\% accuracy drop across the six CNN models including MobileNet, ShuffleNet, and SqueezeNet, which are light-weight networks. Moreover, \textsf{Quantune} achieved 0.59\% higher accuracy for GoogleNet than TensorRT on the NVIDIA GPU. Furthermore, \textsf{Quantune} led to a significant improvement in accuracy by achieving an improvement of 32.52\% on the accelerator compared with the published study on VTA. Finally, by extending the DL compiler stack, we reduced the effort needed to efficiently execute the quantized CNN models on diverse hardware devices.


\section*{Acknowledgments}
This work was supported by Institute of Information \& communications Technology Planning \& Evaluation (IITP) grant funded by the Korea government(MSIT) (No.2018-0-00769, Neuromorphic Computing Software Platform for Artificial Intelligence Systems).


\bibliography{elsarticle-template}

\end{document}